\documentclass[lettersize,journal]{IEEEtran}
\usepackage{amsmath,amsfonts}
\usepackage{array}
\usepackage[caption=false,font=normalsize,labelfont=sf,textfont=sf]{subfig}
\usepackage{textcomp}
\usepackage{stfloats}
\usepackage{url}
\usepackage{verbatim}
\usepackage{graphicx}
\usepackage{cite}
\usepackage{times}
\usepackage{epsfig}
\usepackage{graphicx}
\usepackage{amsmath}
\usepackage{amssymb}
\usepackage{xcolor}
\usepackage{booktabs, makecell, tabularx}
\usepackage{multirow}
\usepackage{dsfont}
\usepackage{enumitem}
\usepackage{algorithm}
\usepackage{algpseudocode}
\usepackage{xspace}
\usepackage{caption,cite}
   
\usepackage{pifont}%

\definecolor{mygray}{gray}{.93}
\definecolor{mygray1}{gray}{.99}

\definecolor{urlcolor}{RGB}{255,105,180}
\definecolor{citecolor}{RGB}{0, 80, 200}
\definecolor{linkcolor}{HTML}{ED1C24}
\definecolor{highlightcolor}{HTML}{ABCDEF}

\usepackage[pagebackref=false,breaklinks=true,letterpaper=true,colorlinks,bookmarks=true]{hyperref}

\newlength\secmargin
\newlength\subsecmargin
\newlength\paramargin
\newlength\figmargin
\newlength\eqmargin
\usepackage[capitalize]{cleveref}
\crefname{section}{Sec.}{Secs.}
\Crefname{section}{Section}{Sections}
\Crefname{table}{Table}{Tables}
\crefname{table}{Tab.}{Tabs.}
\crefname{algorithm}{Algo.}{Algos.}

\usepackage{makecell}
\usepackage{colortbl}
\usepackage{booktabs}
\usepackage{amsmath}
\usepackage{amssymb}
\usepackage{mathtools}
\usepackage{amsthm}
\usepackage{amsmath}
\usepackage{amsfonts}

\usepackage{bm}

\providecommand{\impath}[1]{}
\providecommand{\impatha}[1]{}
\providecommand{\impathb}[1]{}
\providecommand{\impathc}[1]{}
\providecommand{\impathd}[1]{}
\providecommand{\impathe}[1]{}

\newcommand{\figteaser}{
    \vspace{-0.8em}
    \centering
    \includegraphics[width=0.93\textwidth]{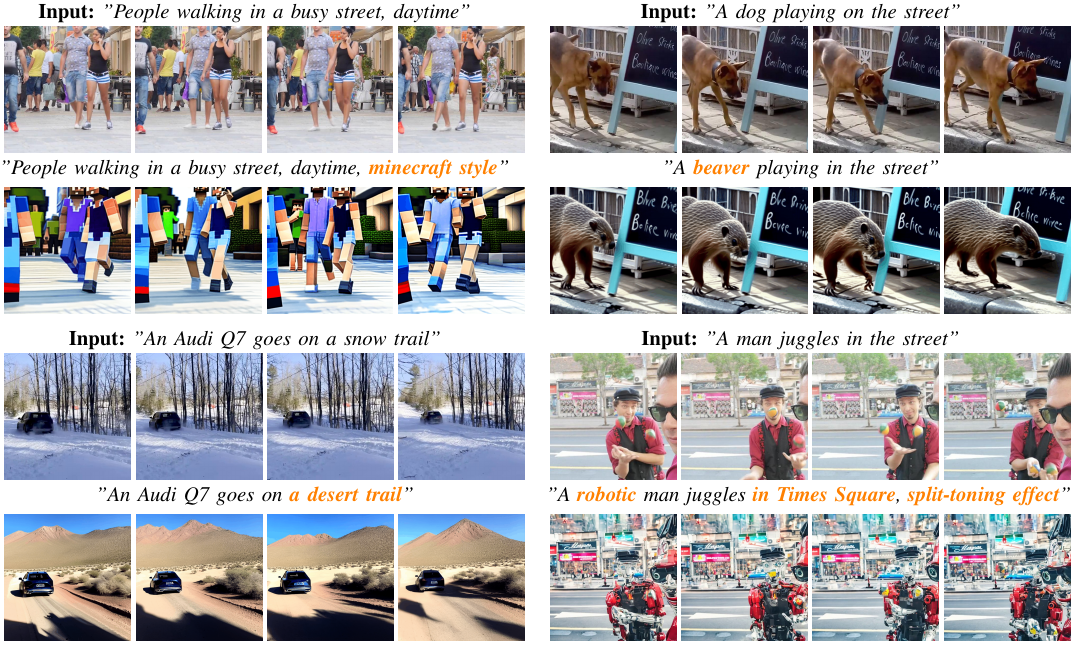}
    \captionof{figure}{FluencyVE is a lightweight and fast video editing method that can efficiently and accurately modify backgrounds, objects, styles and make multiple changes.}
    \label{fig:teaser}
    \vspace{0.8em}
}

\newcommand{\figmethod}{
    \begin{figure*}[t]
      \centerline{\includegraphics[width=\textwidth]{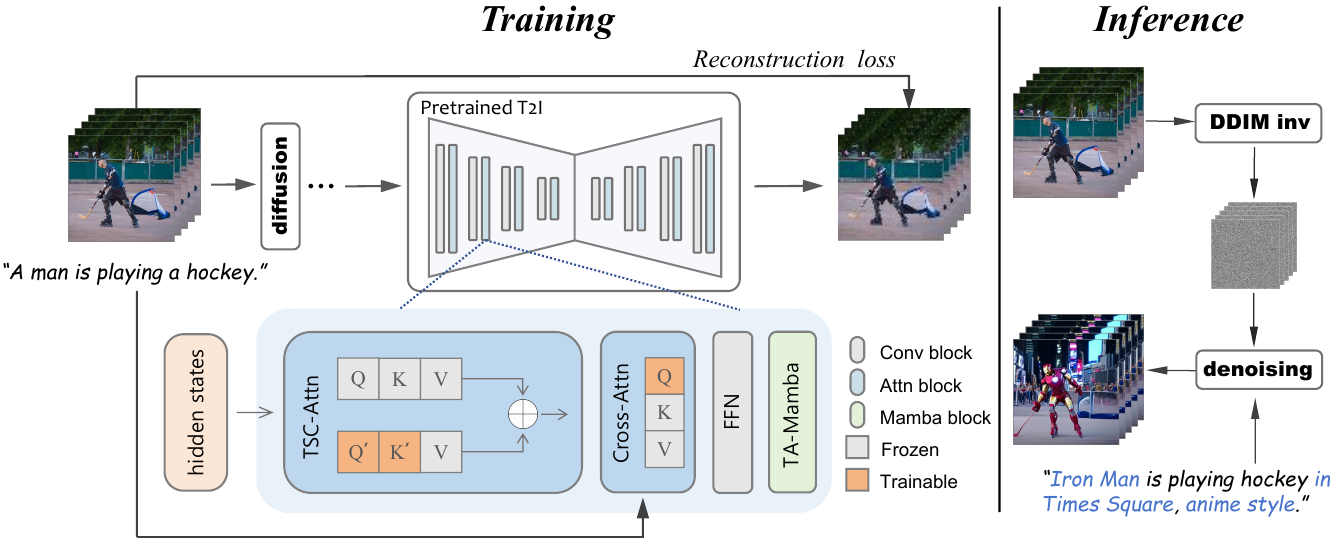}}
      \caption{Illustration of the proposed FluencyVE for one-shot video editing. Given a text-video pair, we approximate the weight matrices \( Q \) and \( K \) in the original sparse-causal attention layer as \( Q^{\prime} \) and \( K^{\prime} \), and calculate new attention scores using a weighted average to reduce parameter count and training cost, subsequently fine-tuning only \( Q^{\prime} \) and \( K^{\prime} \).We also introduce a time-aware linear sequence module, TA-Mamba (Fig. \ref{fig:padding}(b)), to further enhance temporal awareness of video features, enabling smooth and continuous video editing effects. During fine-tuning, we follow the fine-tuning strategy from Tune-A-Video, updating only the weights of \( Q \) in the cross attention layer. During inference, we sample a novel video from the latent noise, which is inverted from the input video and guided by an edited prompt.}
      \label{fig:pipe}
      \vspace{-8mm}
    \end{figure*}
}

\newcommand{\figscan}{
   \begin{figure}[!htb]
    \centering

    \makebox[\columnwidth][l]{ 

        \begin{minipage}[t]{0.40\linewidth}  
            \centering
            \includegraphics[width=1\linewidth]{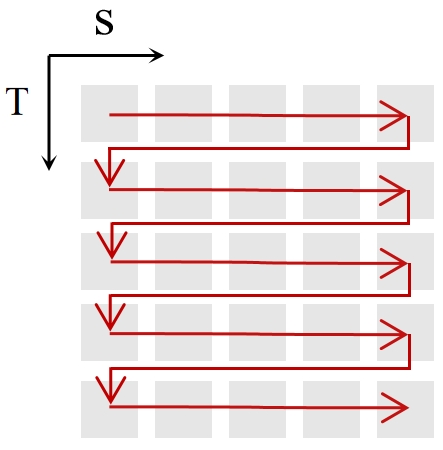}
            \normalsize \parbox{3.5cm}{\centering \hspace{0.5cm} (a) \textit{Spatial forward}, \\ \hspace{0.8cm} \textit{Temporal forward}} 
        \end{minipage}
        \hspace*{0.01\linewidth} 

        \begin{minipage}[t]{0.40\linewidth} 
            \centering
            \includegraphics[width=1\linewidth]{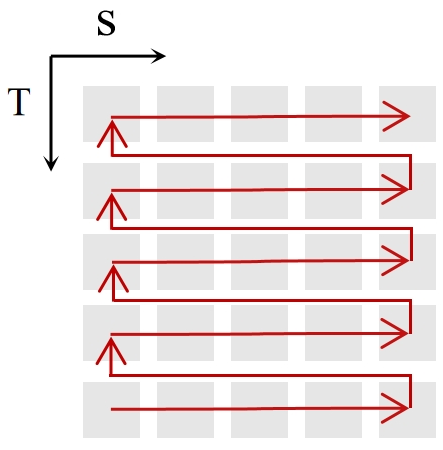}
            \normalsize \parbox{3.5cm}{\centering \hspace{0.5cm} (b) \textit{Spatial forward}, \\ \hspace{0.8cm} \textit{Temporal reverse}} 
        \end{minipage}

    }

    \makebox[\columnwidth][l]{ 

        \begin{minipage}[t]{0.40\linewidth} 
            \centering
            \includegraphics[width=1\linewidth]{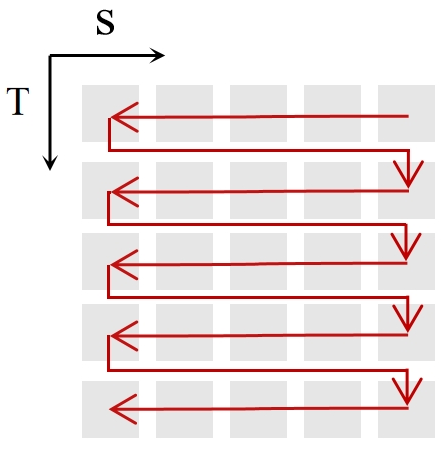}
            \normalsize \parbox{3.5cm}{\centering \hspace{0.5cm} (c) \textit{Spatial reverse}, \\ \hspace{0.8cm} \textit{Temporal forward}}  
        \end{minipage}
        \hspace*{0.01\linewidth}  

        \begin{minipage}[t]{0.40\linewidth} 
            \centering
            \includegraphics[width=1\linewidth]{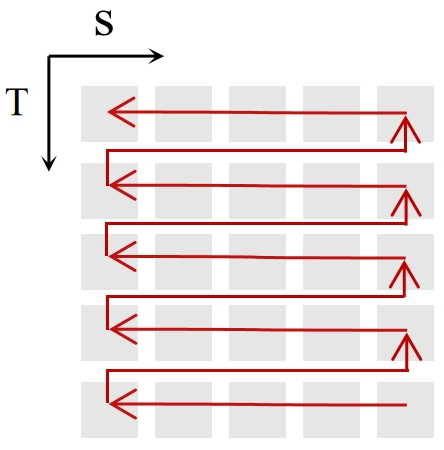}
            \normalsize \parbox{3.5cm}{\centering \hspace{0.5cm} (d) \textit{Spatial reverse}, \\ \hspace{0.8cm} \textit{Temporal reverse}} 
        \end{minipage}

    }
    \caption{Different Scan Methods. Following the Spatial-First rule, we introduce four novel scanning methods by reversing temporal or spatial ordering.}
    \vspace{-2mm}
    \label{fig:scan}
\end{figure}
}

\newcommand{\figpadding}{
\begin{figure*}[t!]
    \centering
    \includegraphics[width=1.0\linewidth]{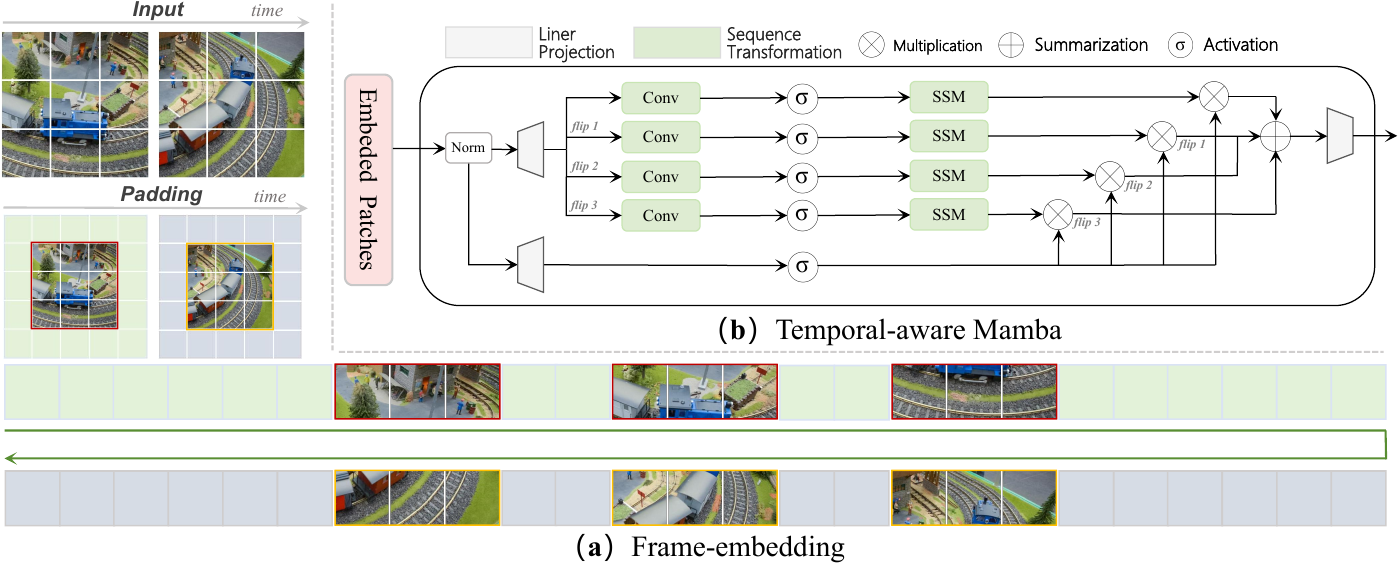}

    \vspace{-0.15cm}  
    \caption{Illustration of the proposed Temporal-aware Mamba. (a) Unique trainable embedding vectors are assigned to each frame, enabling the model to better capture the temporal characteristics and intra-frame distributions. The sequence is then fed as the spatial-temporal forward input into (b) the temporal-aware Mamba , where flip operations generate inputs that follow the four-directional scanning strategy shown in Fig. \ref{fig:scan}, which are then processed by the SSM.}
    \label{fig:padding}
    \vspace{-4mm}
\end{figure*}
}

\newcommand{\figfinetune}{
\begin{figure}[!htb]
    \centering

    \makebox[\columnwidth][l]{

        \begin{minipage}[t]{1\linewidth} 
            \centering
            \includegraphics[width=1\linewidth]{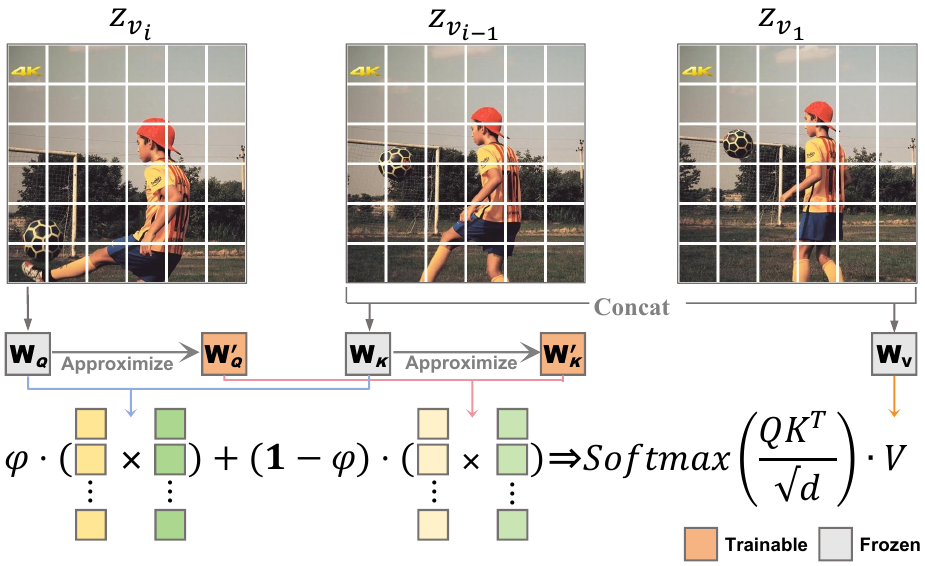}
        \end{minipage}

    }

    \caption{Illustration of the Bypass Attention. Latent features of frame \(v_{i}\), previous frames \(v_{i-1}\) and \(v_{1}\) are projected to the weight matrices, query \(W_{Q}\), key \(W_{K}\) and value \(W_{V}\). We substitute the low-rank approximation matrices \( W_q \) and \( W_k \) for \( W_{Q} \) and \( W_{K} \), respectively, and compute the attention scores through weighted averaging.
}
    \label{fig:finetune}
    \vspace{-4mm}
\end{figure}
}

\newcommand{\figgoodcase}{
    \begin{figure*}[p]
    \centering

    \makebox[0.116\textwidth]{[Source Prompt] A cat in the grass in the sun.} \\
    \includegraphics[width=0.116\textwidth]{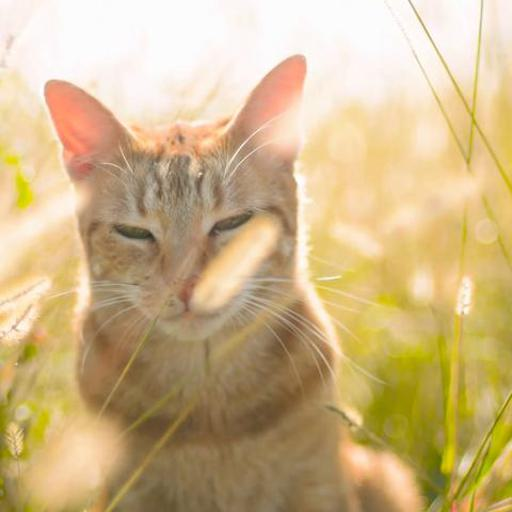}
    \includegraphics[width=0.116\textwidth]{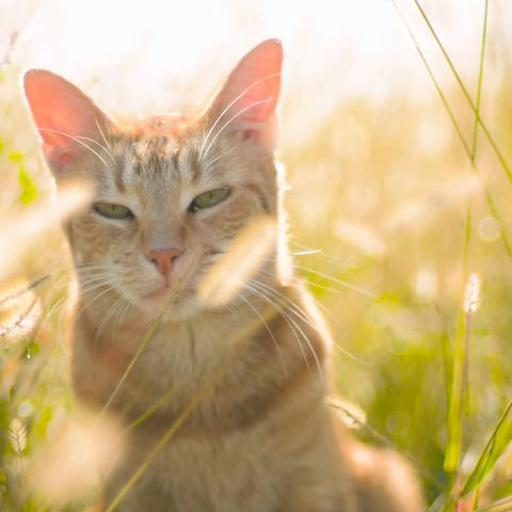}
    \includegraphics[width=0.116\textwidth]{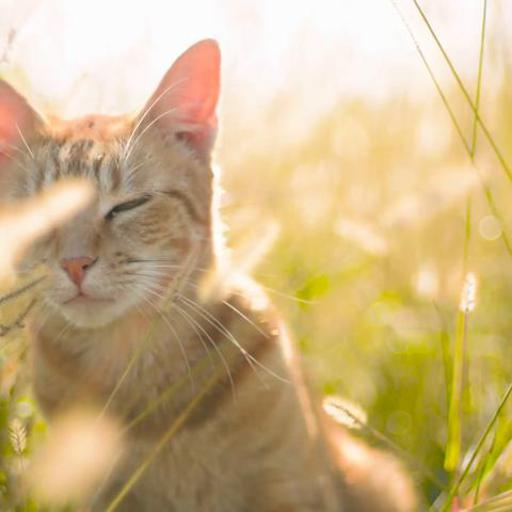}
    \includegraphics[width=0.116\textwidth]{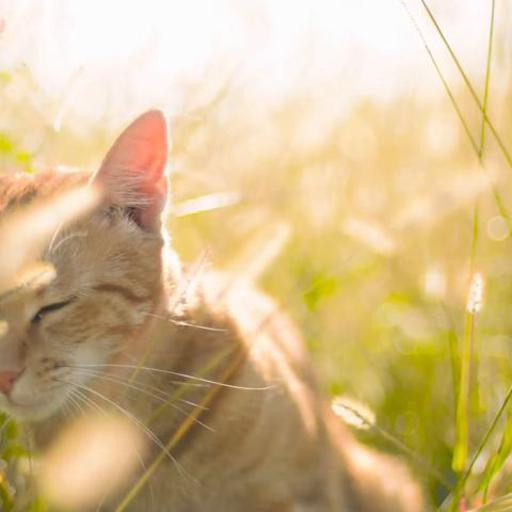}
    \includegraphics[width=0.116\textwidth]{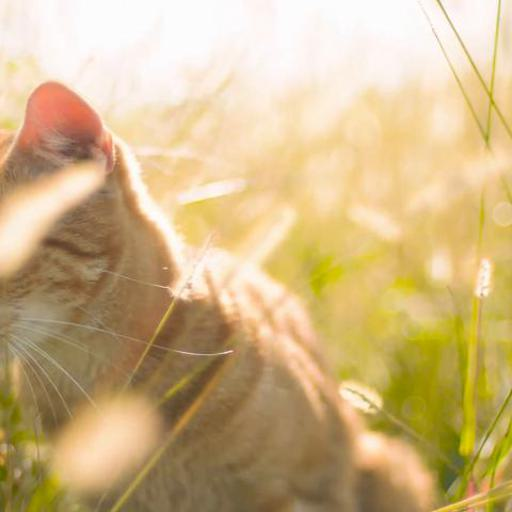}
    \includegraphics[width=0.116\textwidth]{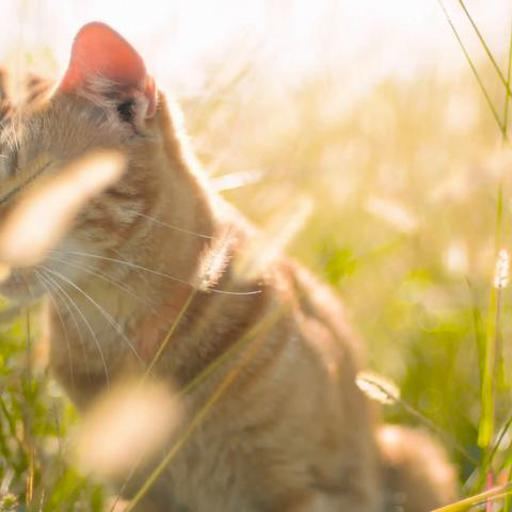}
    \includegraphics[width=0.116\textwidth]{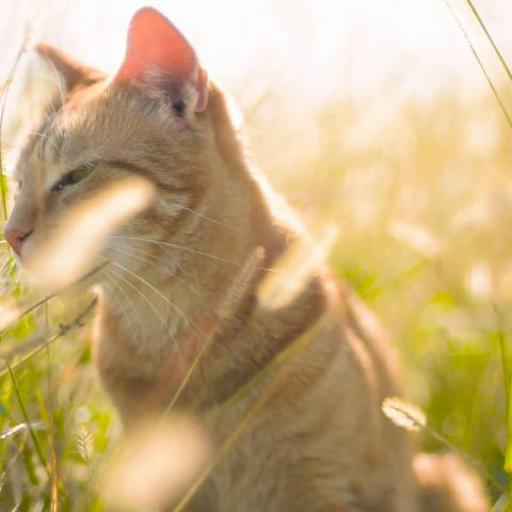}
    \includegraphics[width=0.116\textwidth]{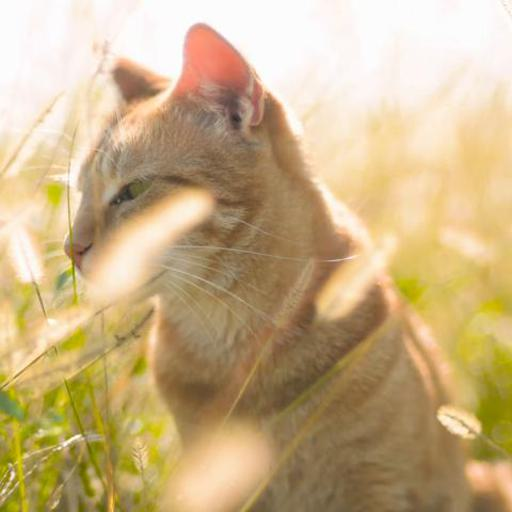}

    \makebox[0.116\textwidth]{A cat \textcolor{magenta}{\textbf{on a beach}} in the sun.} \\
    \includegraphics[width=0.116\textwidth]{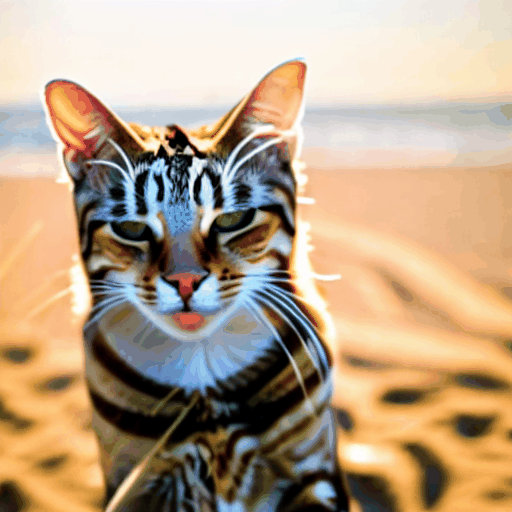}
    \includegraphics[width=0.116\textwidth]{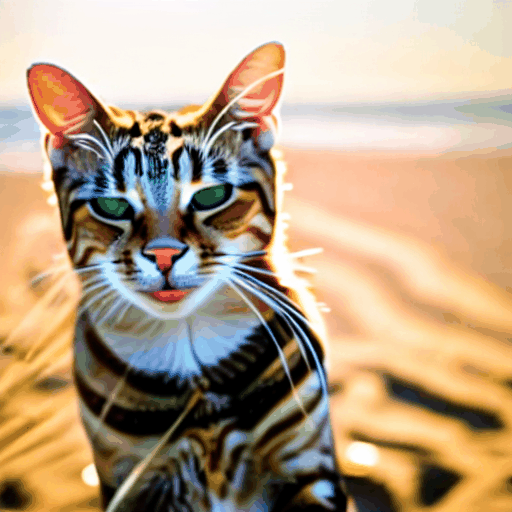}
    \includegraphics[width=0.116\textwidth]{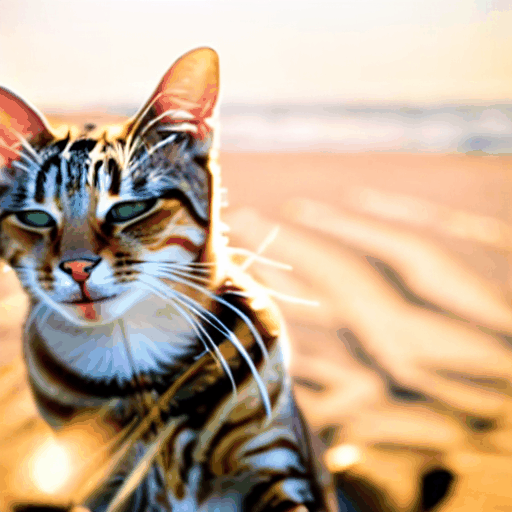}
    \includegraphics[width=0.116\textwidth]{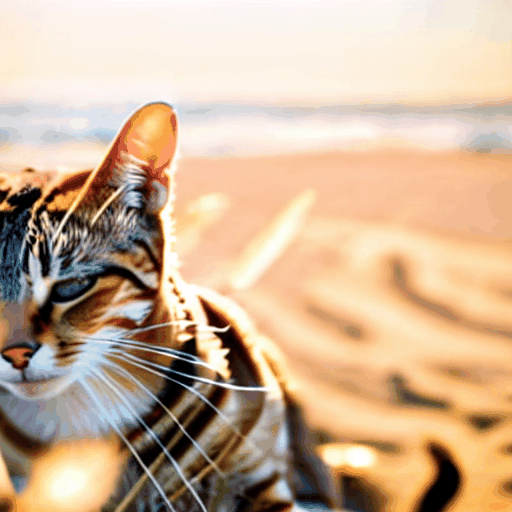}
    \includegraphics[width=0.116\textwidth]{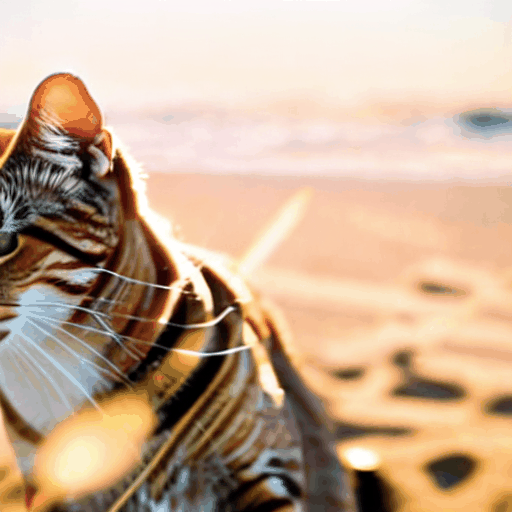}
    \includegraphics[width=0.116\textwidth]{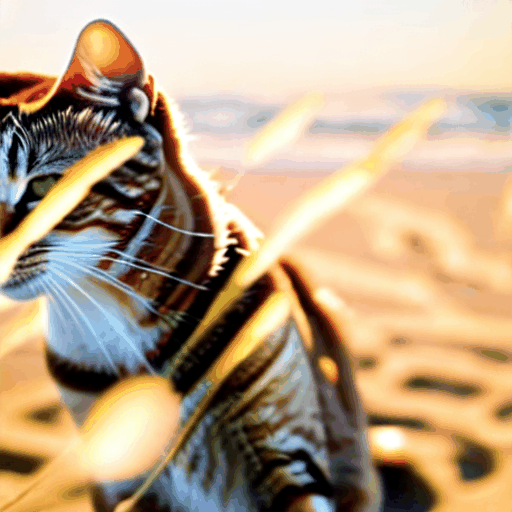}
    \includegraphics[width=0.116\textwidth]{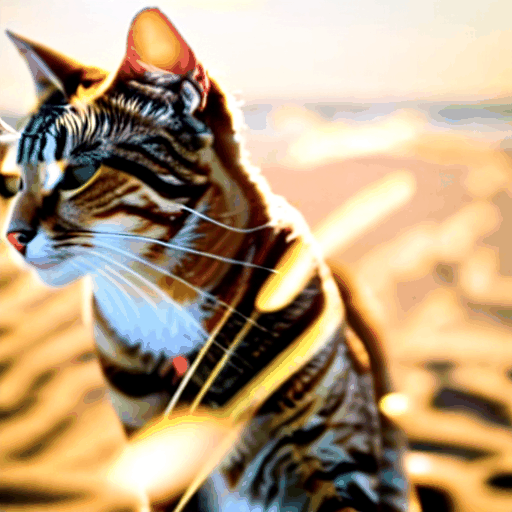}
    \includegraphics[width=0.116\textwidth]{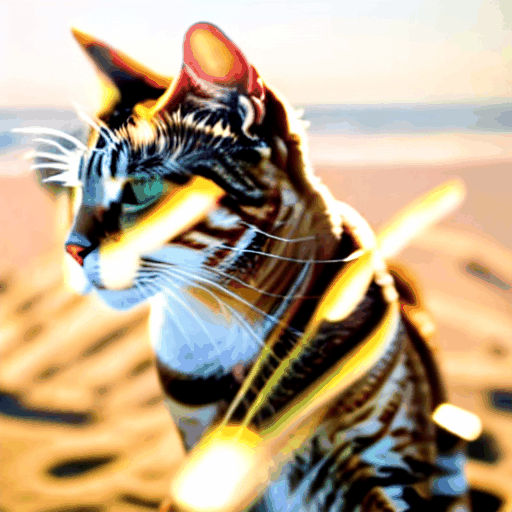}
    
    \makebox[0.116\textwidth]{\textcolor{magenta}{\textbf{A dog}} in the grass in the sun.} \\
    \includegraphics[width=0.116\textwidth]{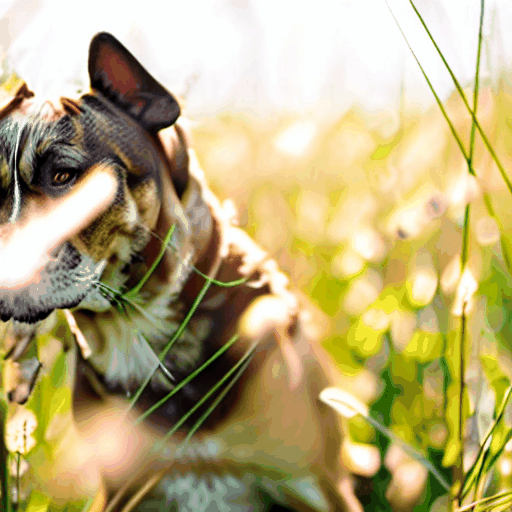}
    \includegraphics[width=0.116\textwidth]{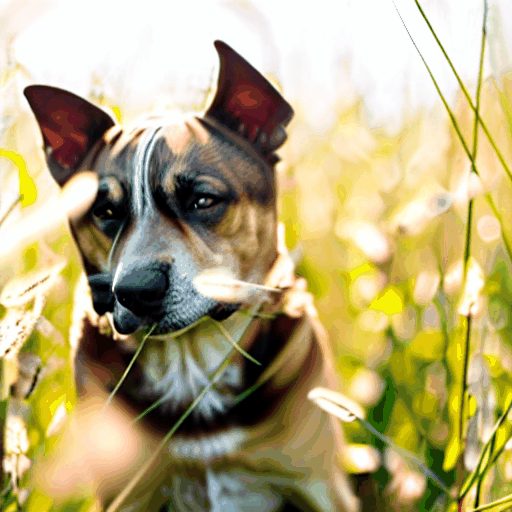}
    \includegraphics[width=0.116\textwidth]{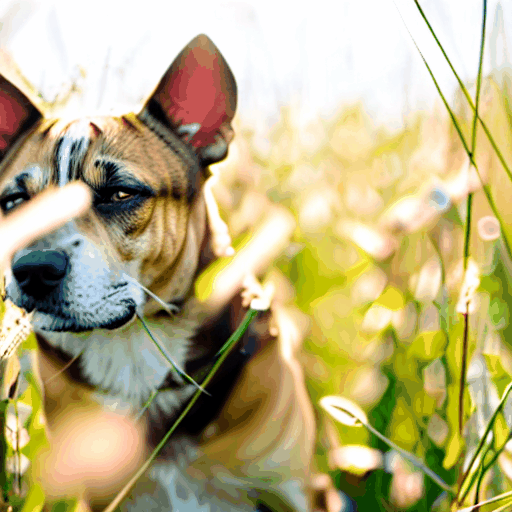}
    \includegraphics[width=0.116\textwidth]{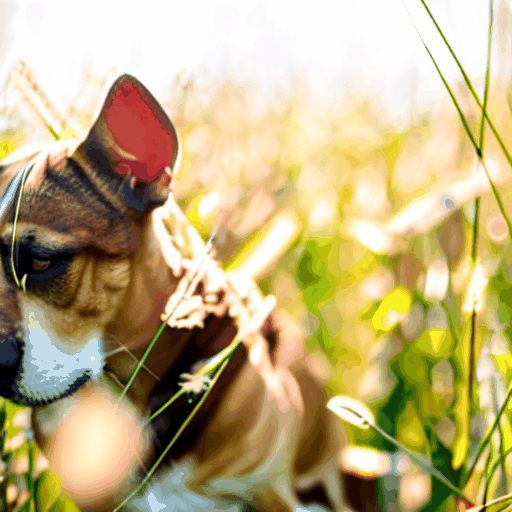}
    \includegraphics[width=0.116\textwidth]{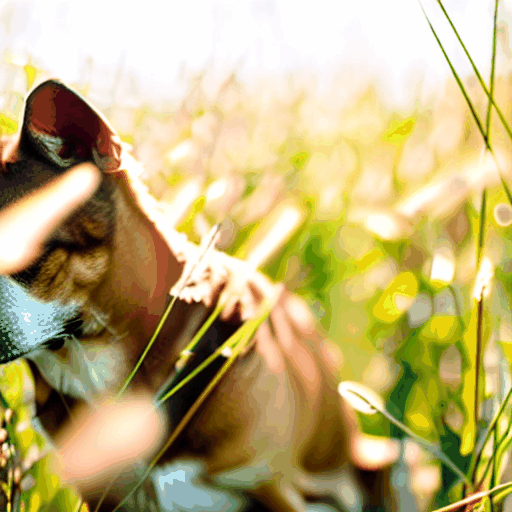}
    \includegraphics[width=0.116\textwidth]{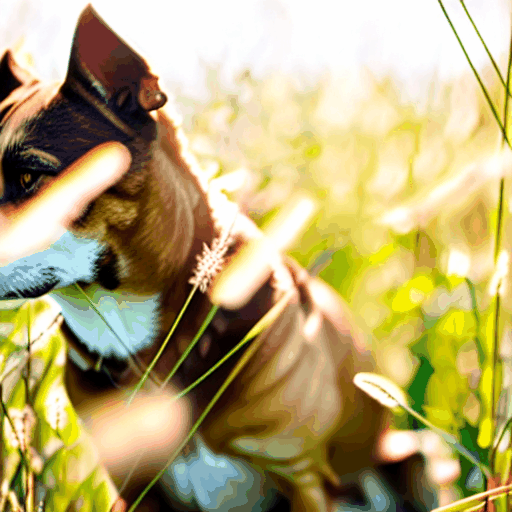}
    \includegraphics[width=0.116\textwidth]{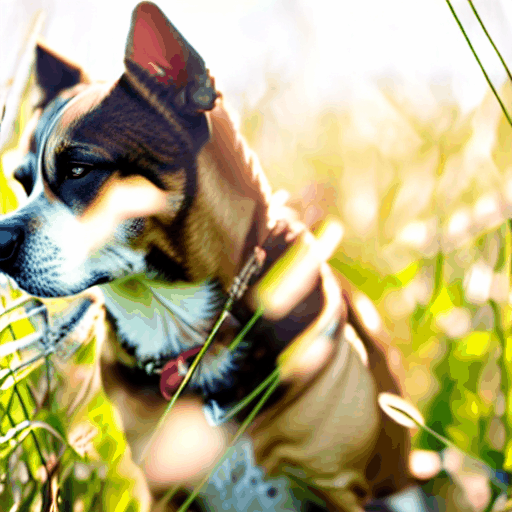}
    \includegraphics[width=0.116\textwidth]{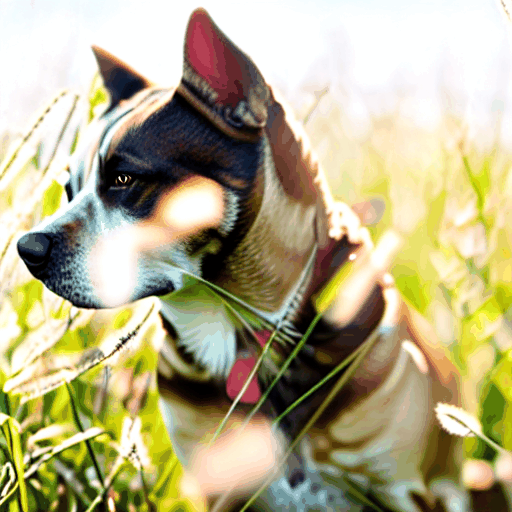}    

    \makebox[0.116\textwidth]{\textcolor{magenta}{\textbf{A lion}} in the grass in the sun, \textcolor{magenta}{\textbf{surrounded by butterflies}}.} \\
    \includegraphics[width=0.116\textwidth]{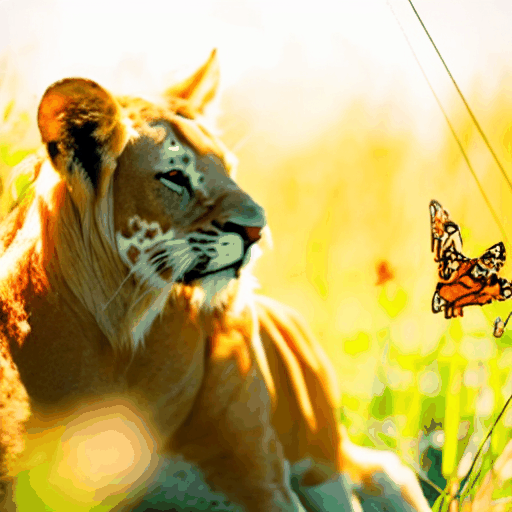}
    \includegraphics[width=0.116\textwidth]{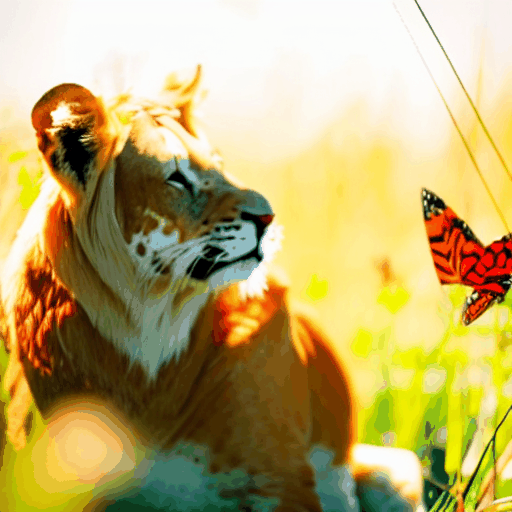}
    \includegraphics[width=0.116\textwidth]{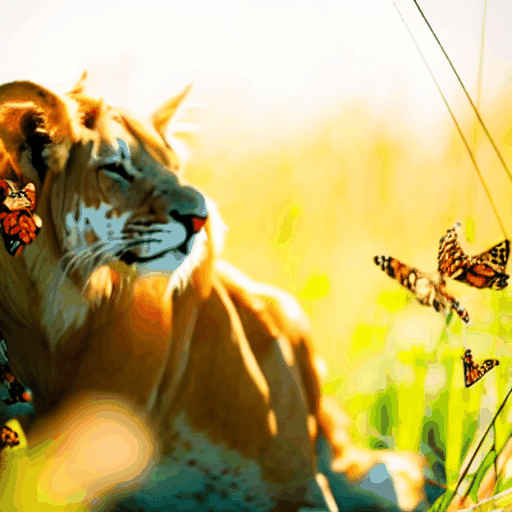}
    \includegraphics[width=0.116\textwidth]{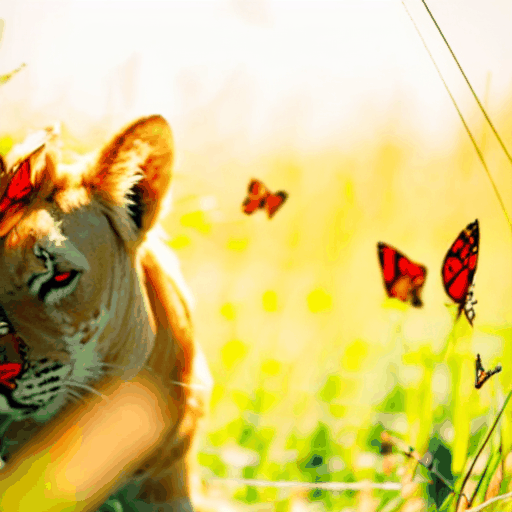}
    \includegraphics[width=0.116\textwidth]{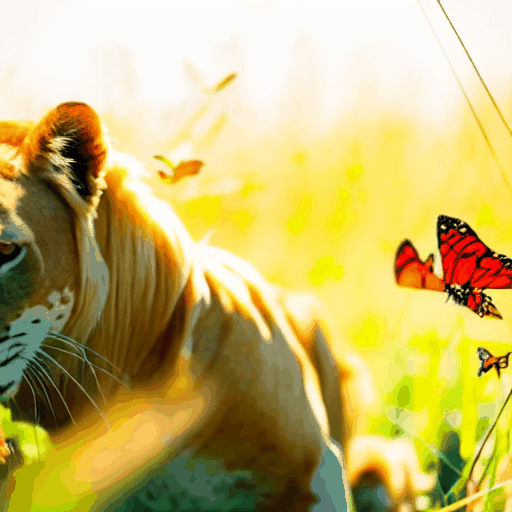}
    \includegraphics[width=0.116\textwidth]{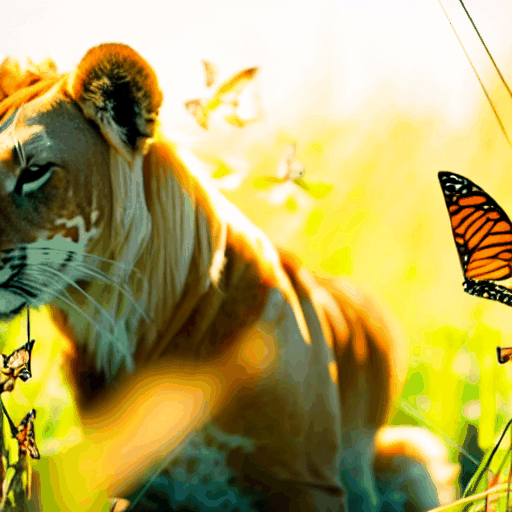}
    \includegraphics[width=0.116\textwidth]{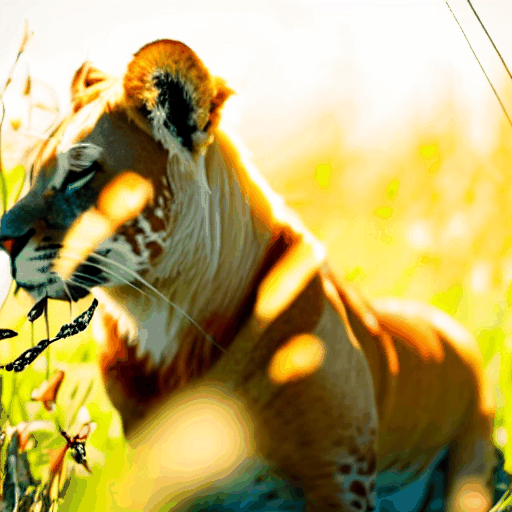}
    \includegraphics[width=0.116\textwidth]{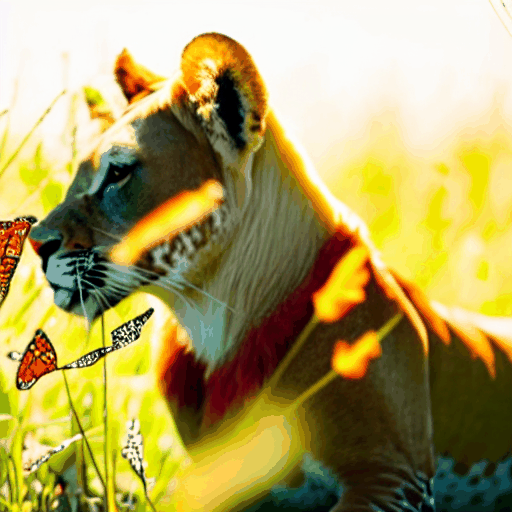}
    
    \makebox[0.116\textwidth]{} 

    \makebox[0.116\textwidth]{[Source Prompt] A jeep car is moving on the road.} \\
    \includegraphics[width=0.116\textwidth]{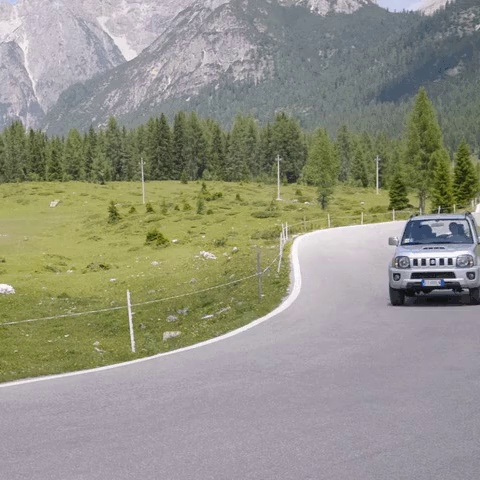}
    \includegraphics[width=0.116\textwidth]{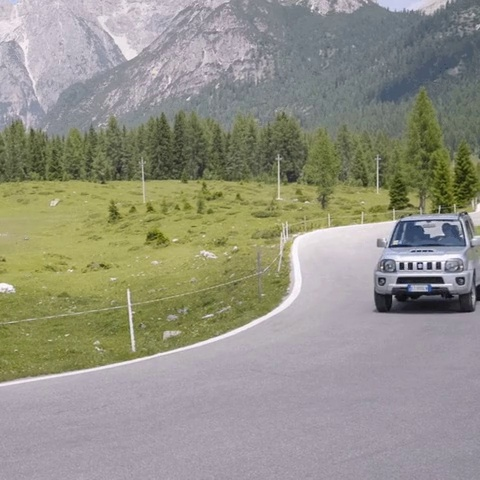}
    \includegraphics[width=0.116\textwidth]{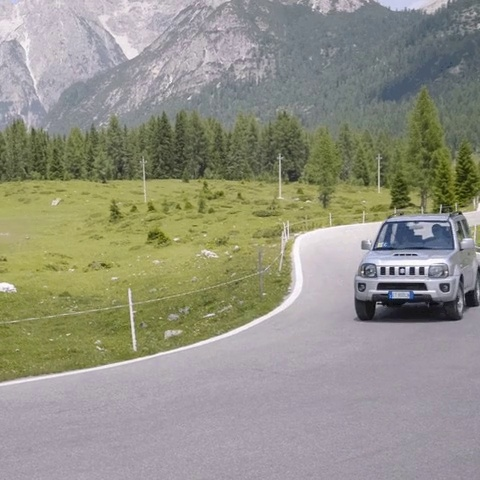}
    \includegraphics[width=0.116\textwidth]{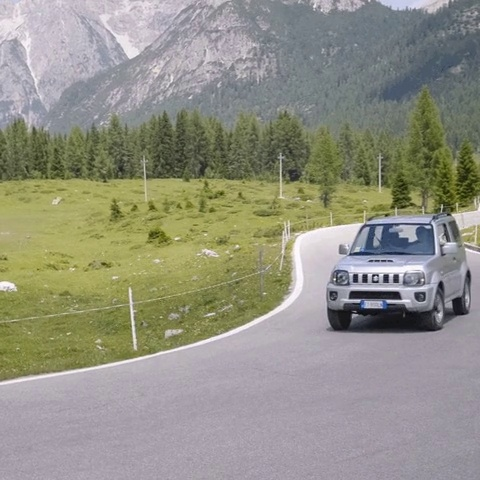}
    \includegraphics[width=0.116\textwidth]{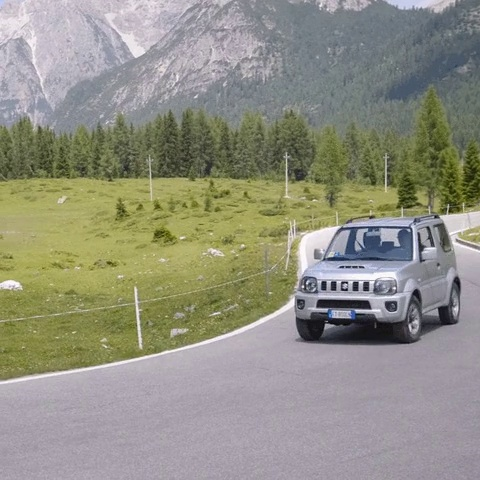}
    \includegraphics[width=0.116\textwidth]{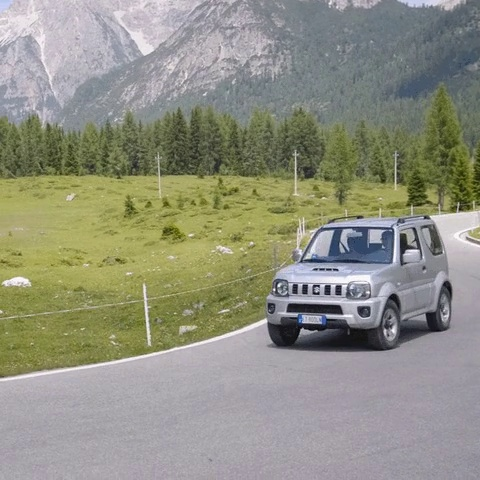}
    \includegraphics[width=0.116\textwidth]{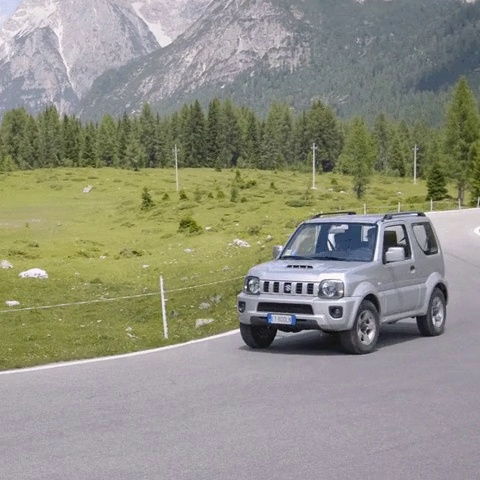}
    \includegraphics[width=0.116\textwidth]{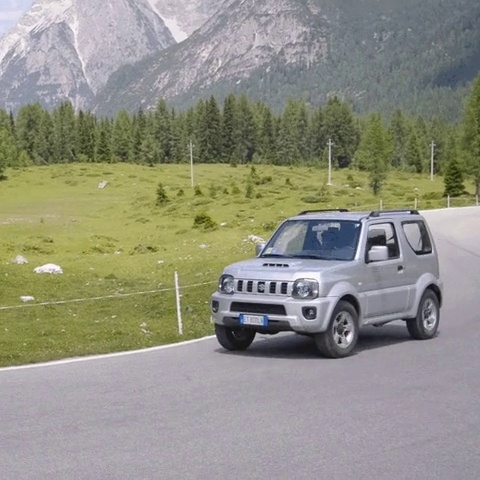}

    \makebox[0.116\textwidth]{A jeep car is moving \textcolor{magenta}{\textbf{on the snow}}.} \\
    \includegraphics[width=0.116\textwidth]{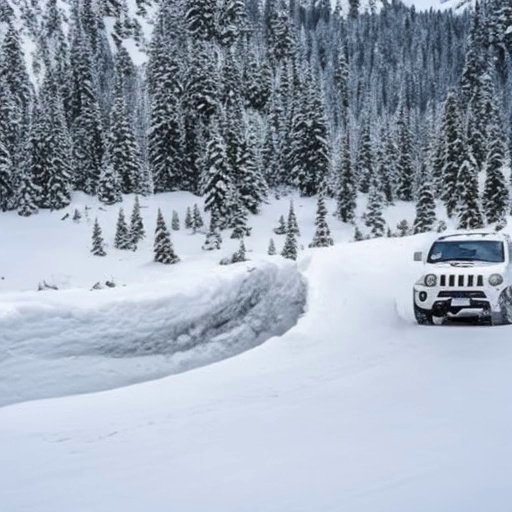}
    \includegraphics[width=0.116\textwidth]{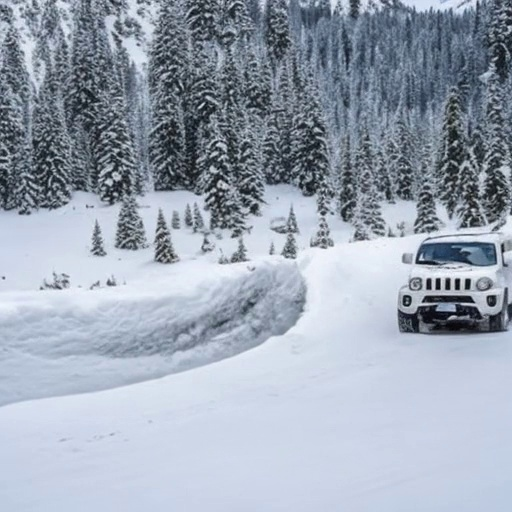}
    \includegraphics[width=0.116\textwidth]{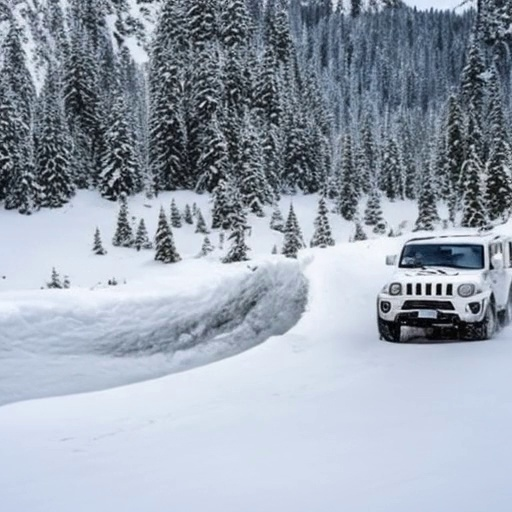}
    \includegraphics[width=0.116\textwidth]{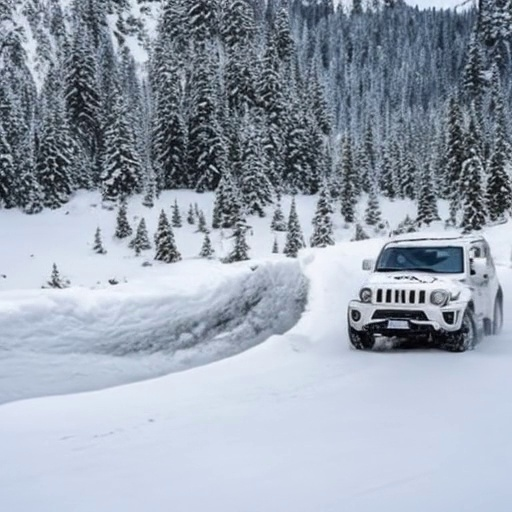}
    \includegraphics[width=0.116\textwidth]{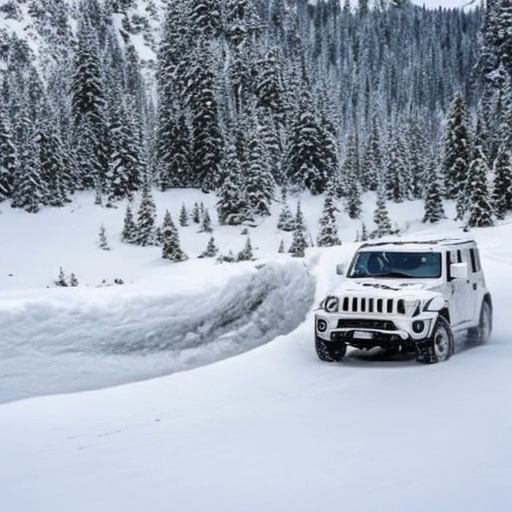}
    \includegraphics[width=0.116\textwidth]{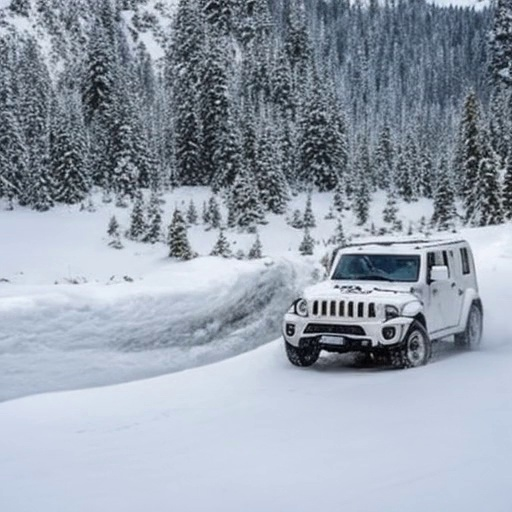}
    \includegraphics[width=0.116\textwidth]{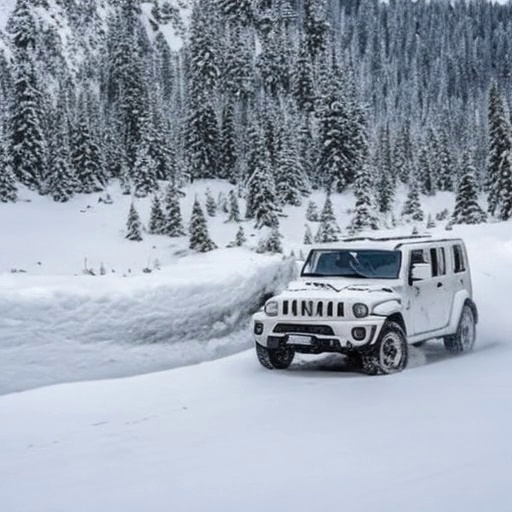}
    \includegraphics[width=0.116\textwidth]{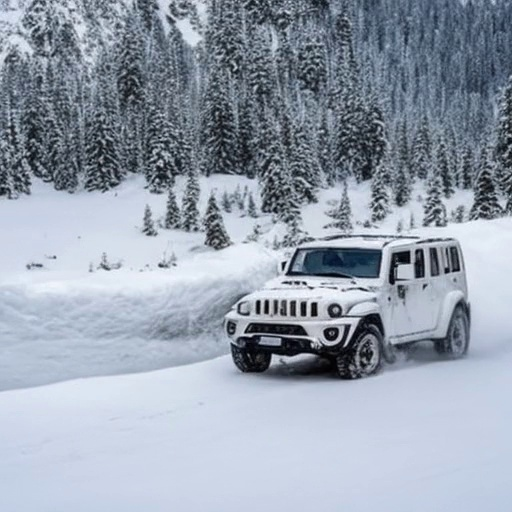}
    
    \makebox[0.116\textwidth]{\textcolor{magenta}{\textbf{A sports car}} is moving on the road.} \\
    \includegraphics[width=0.116\textwidth]{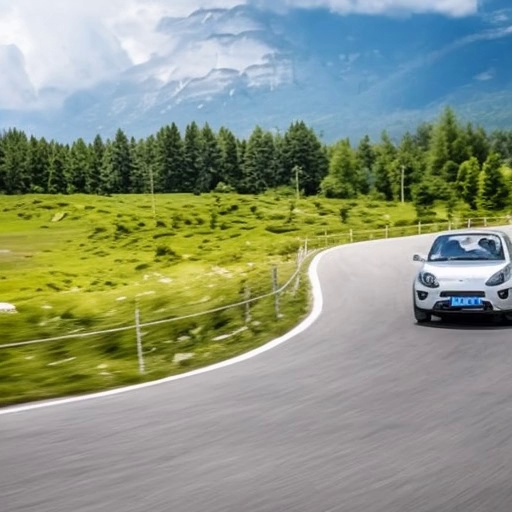}
    \includegraphics[width=0.116\textwidth]{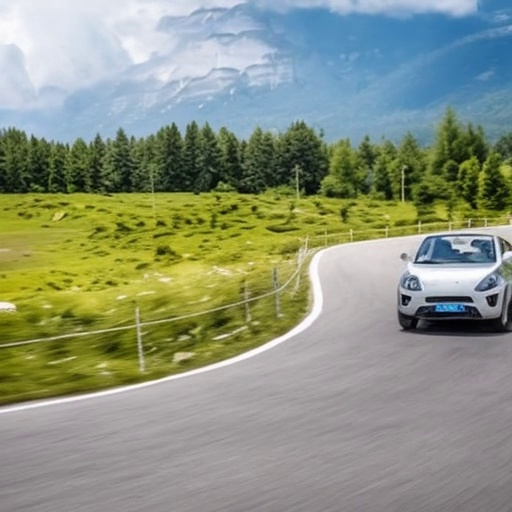}
    \includegraphics[width=0.116\textwidth]{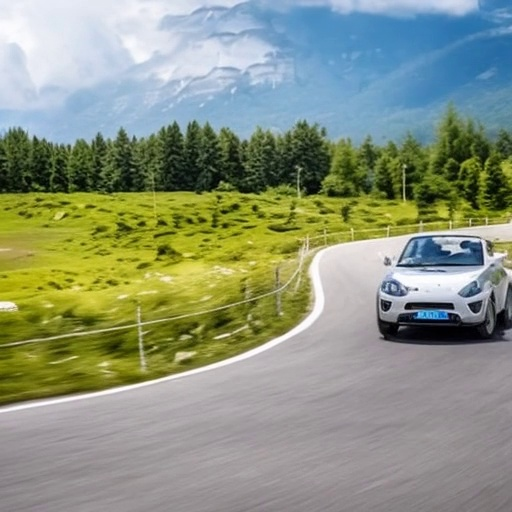}
    \includegraphics[width=0.116\textwidth]{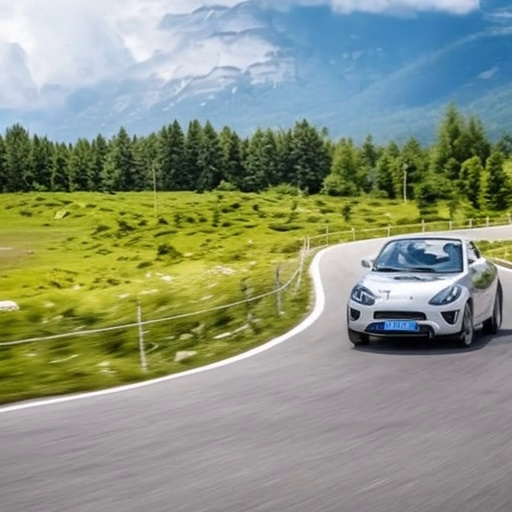}
    \includegraphics[width=0.116\textwidth]{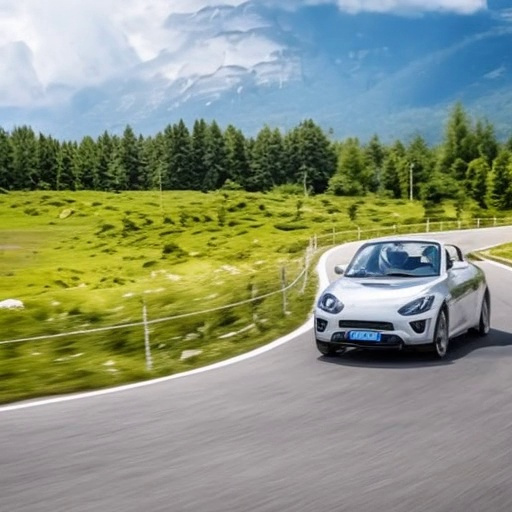}
    \includegraphics[width=0.116\textwidth]{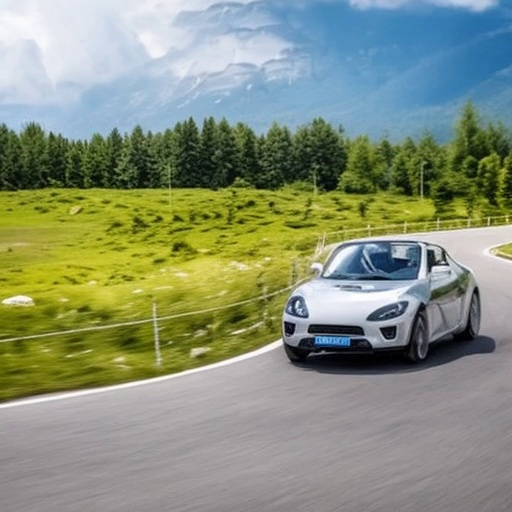}
    \includegraphics[width=0.116\textwidth]{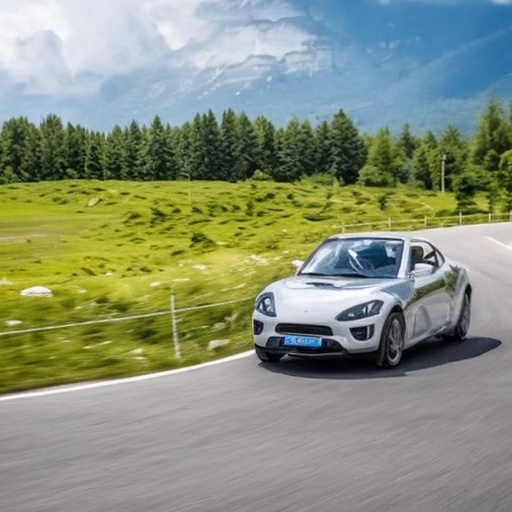}
    \includegraphics[width=0.116\textwidth]{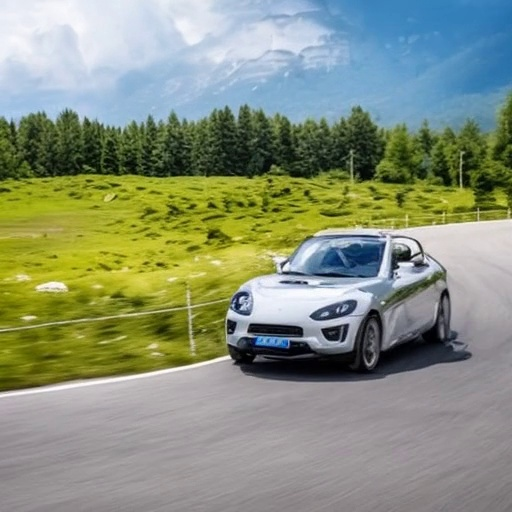}    

    \makebox[0.116\textwidth]{\textcolor{magenta}{\textbf{An AE86}} is moving on the road, \textcolor{magenta}{\textbf{cartoon style}}.} \\
    \includegraphics[width=0.116\textwidth]{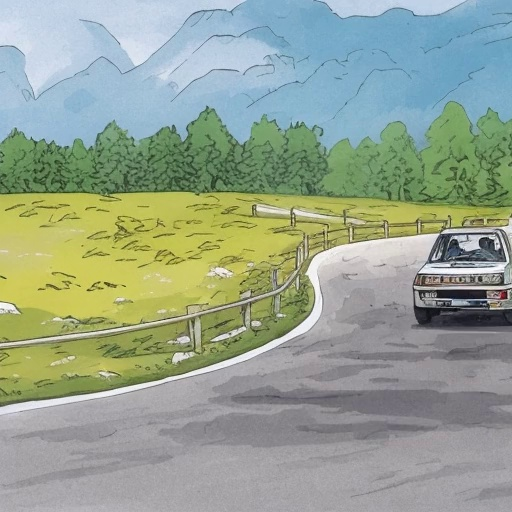}
    \includegraphics[width=0.116\textwidth]{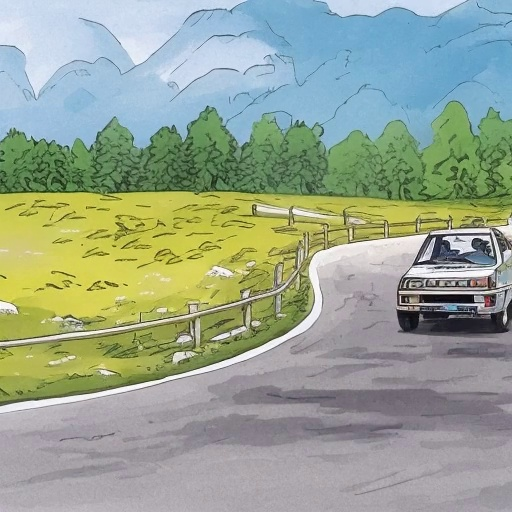}
    \includegraphics[width=0.116\textwidth]{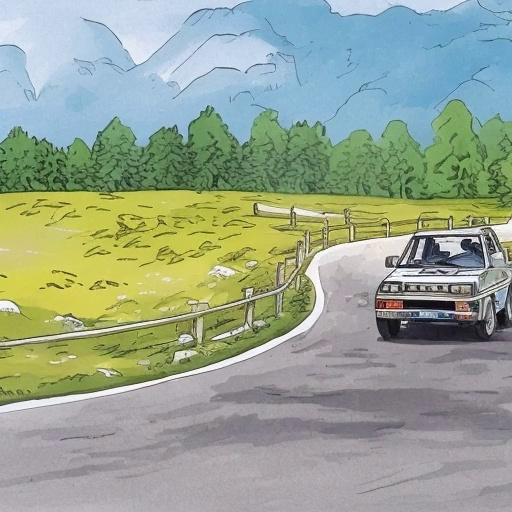}
    \includegraphics[width=0.116\textwidth]{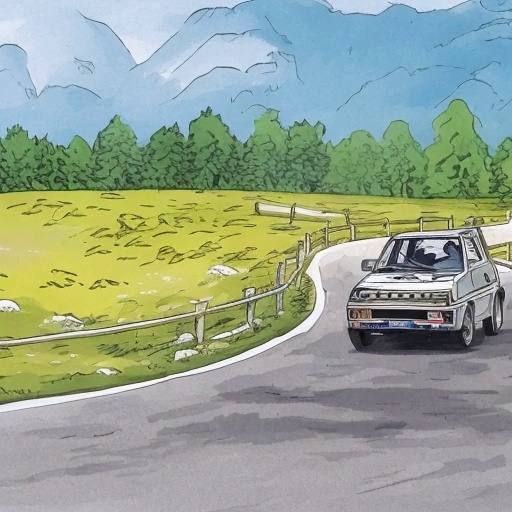}
    \includegraphics[width=0.116\textwidth]{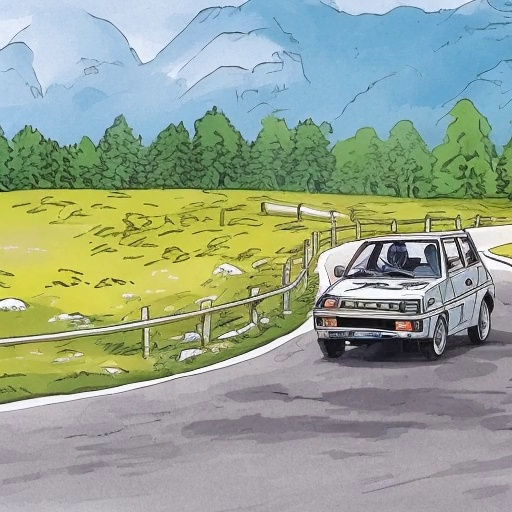}
    \includegraphics[width=0.116\textwidth]{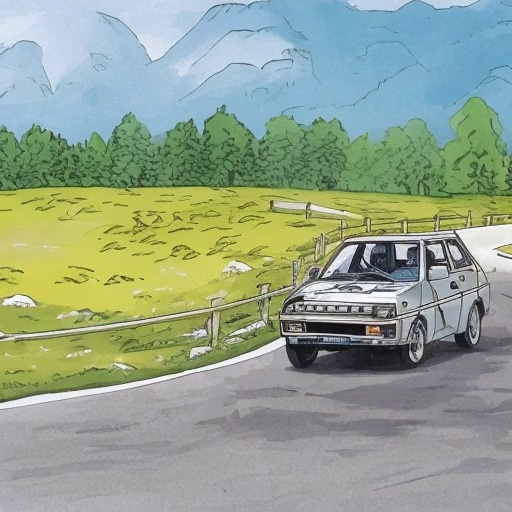}
    \includegraphics[width=0.116\textwidth]{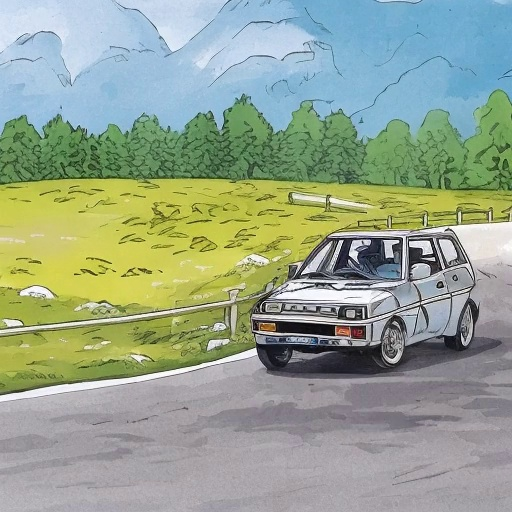}
    \includegraphics[width=0.116\textwidth]{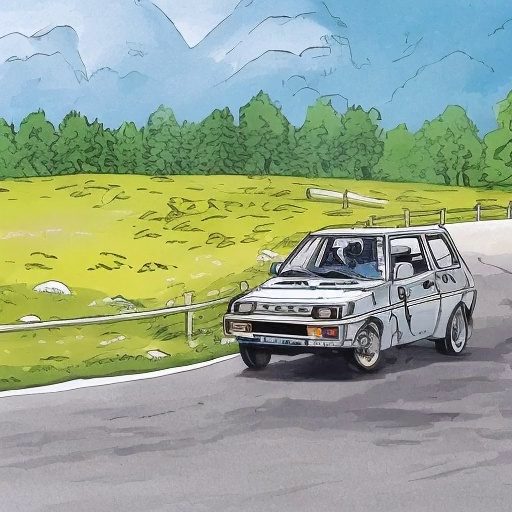}
        
    \caption{Video editing results from various input videos and prompts. Our model produces temporally consistent videos that accurately follow text prompts while preserving the original frame structure. }
    \label{fig:goodcase}
    \end{figure*}
}

\newcommand{\figcomparison}{
\begin{figure}[t!]
    \centering
    \includegraphics[width=1.0\linewidth]{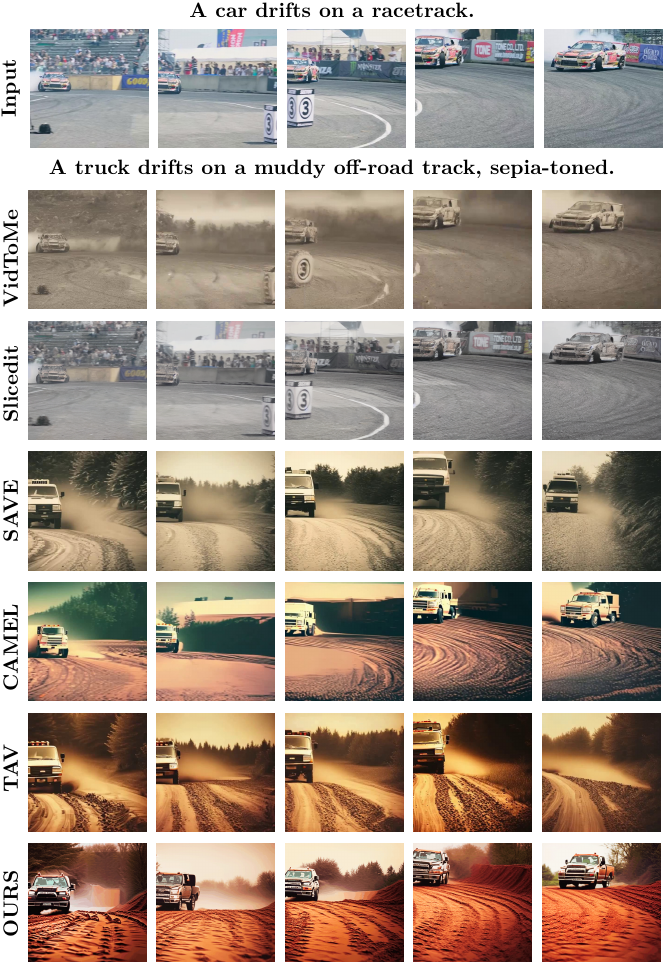}

    \vspace{-0.15cm}  
    \caption{Qualitative comparison with other methods. Our model achieves temporal consistency and fidelity with the input video, preserving both the style coherence and frame structure effciently.}
    \label{fig:comparison}
    \vspace{-6mm}
\end{figure}
}

\newcommand{\figabaa}{
\begin{figure}[htbp]
	\centering
		\includegraphics[width=1.0\linewidth]{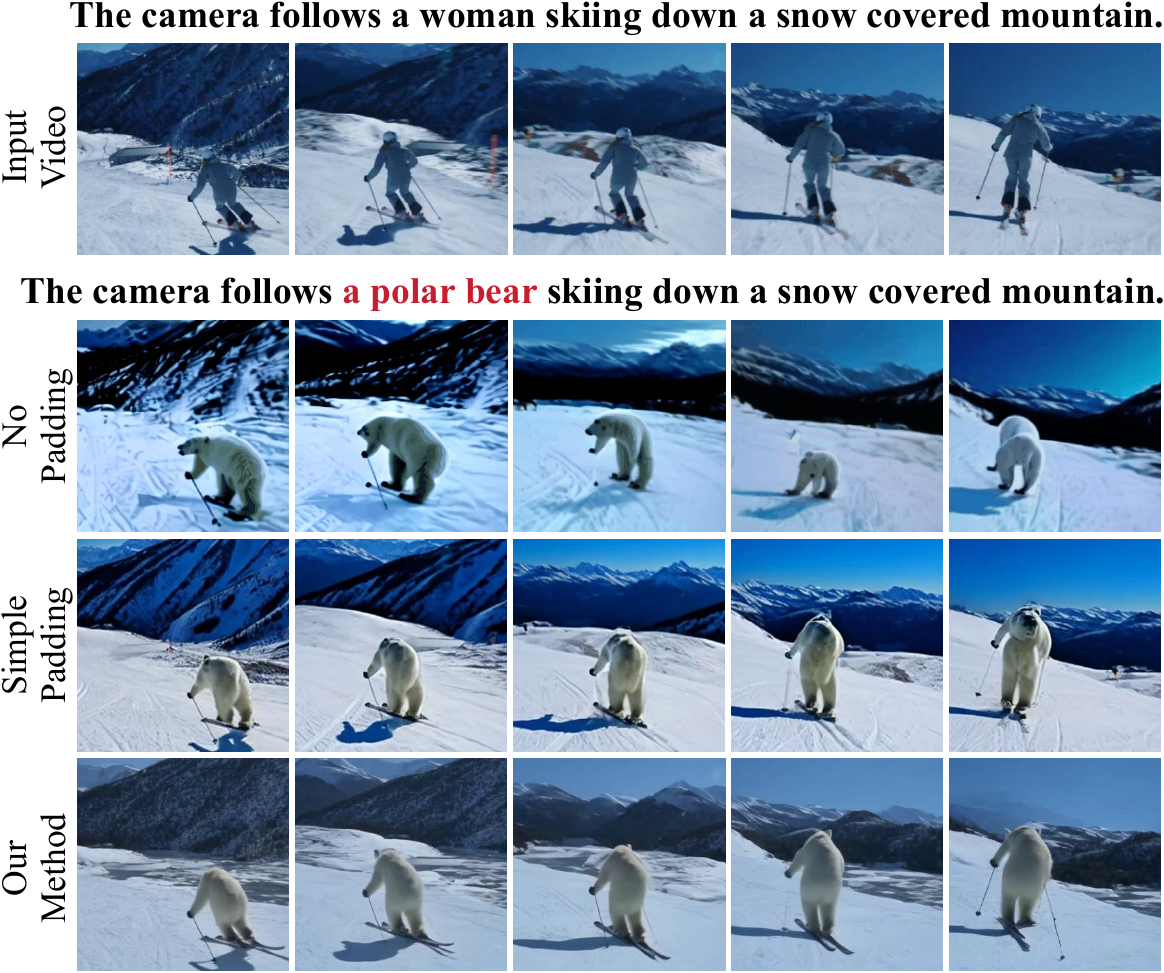}
	\centering
	\caption{Ablations on the effectiveness of the Padding Strategy.}
	\label{fig:aba_1}
    \vspace{-6mm}
\end{figure}
}

\newcommand{\figabab}{
\begin{figure}[t!]
	\centering
		\includegraphics[width=1.0\linewidth]{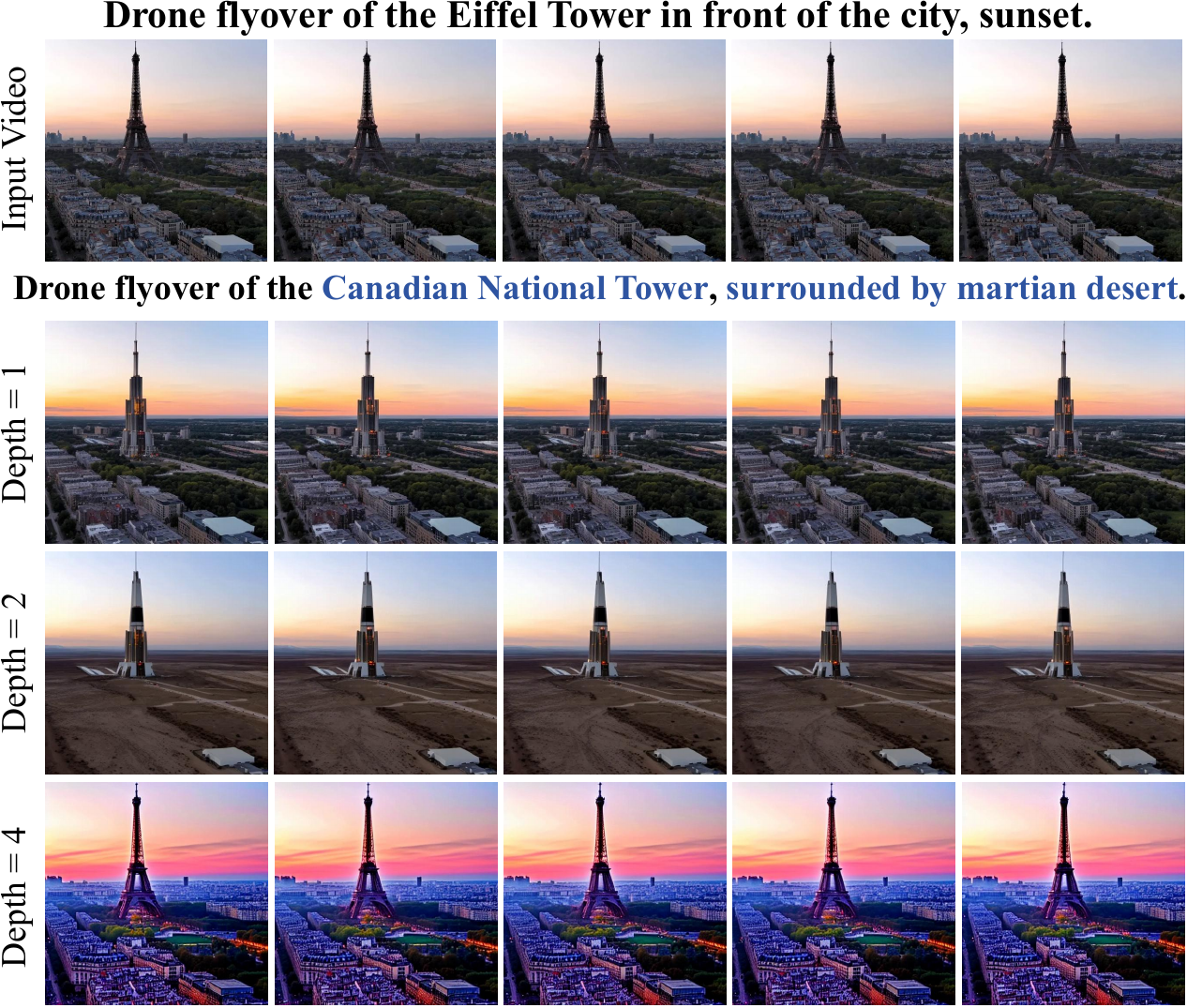}
	\centering
	\caption{Ablations on the depth of the Temporal-aware Mamba block.}
	\label{fig:aba_2}
    \vspace{-6mm}
\end{figure}
}

\newcommand{\figabac}{
\begin{figure}[t!]
	\centering
		\includegraphics[width=1.0\linewidth]{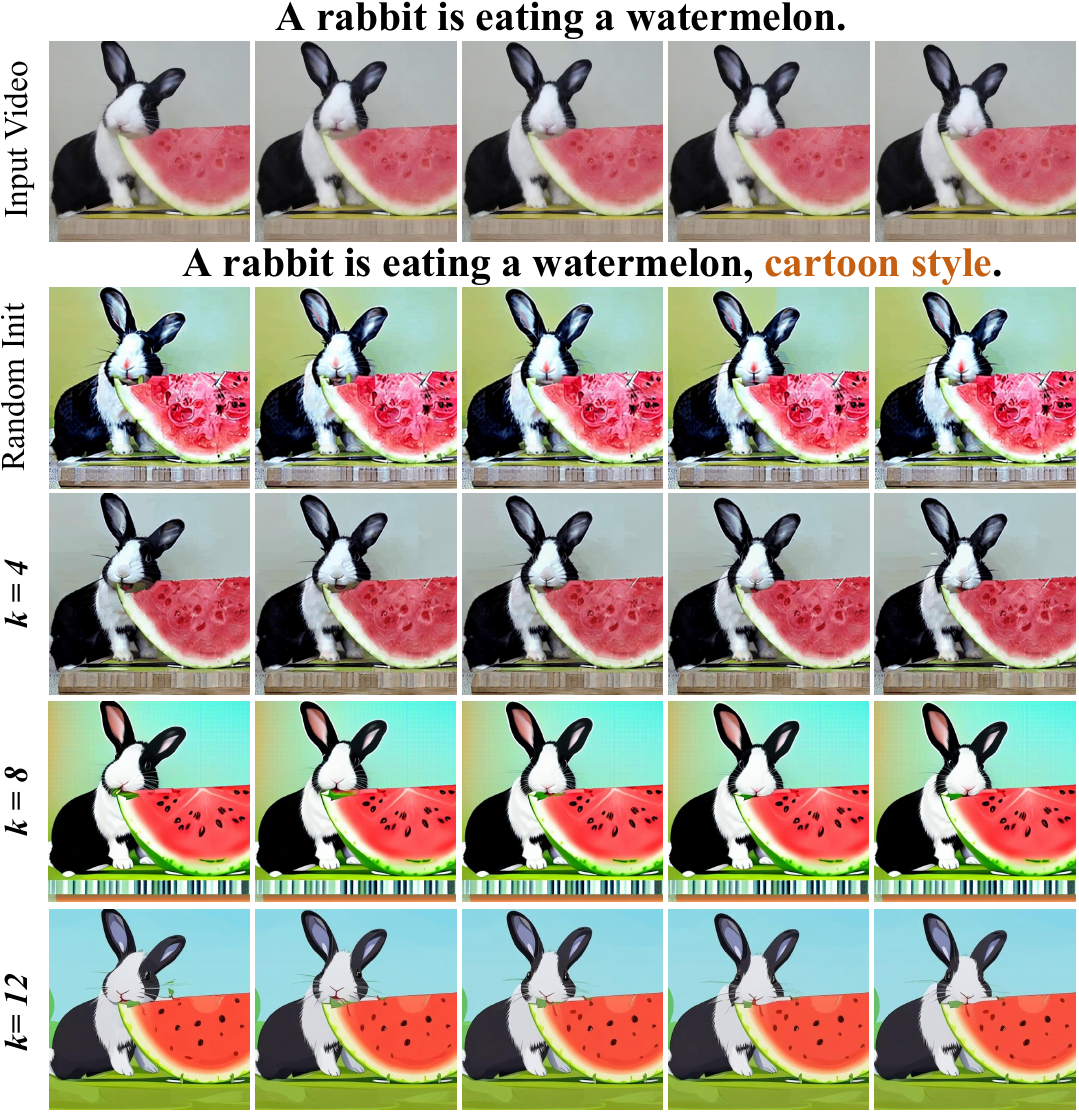}
	\centering
	\caption{Ablations on the \(k\) value of the low-rank approximation matrix. }
	\label{fig:aba_3}
    \vspace{-8mm}
\end{figure}
}

\usepackage{blindtext}
\title{here title}
\let\oldtwocolumn\twocolumn

\renewcommand\twocolumn[1][]{%
    \oldtwocolumn[{#1}{
    \figteaser
    }]
}

\begin{document}
\title{FluencyVE: Marrying Temporal-Aware Mamba with Bypass Attention for Video Editing}

\author{{
Mingshu~Cai, 
Yixuan Li,
Osamu Yoshie and~Yuya Ieiri}
\thanks{Mingshu~Cai, Osamu Yoshie, and Yuya Ieiri are with
Waseda University, 
  Japan (e-mail: mignshucai@fuji.waseda.jp
  yoshie@waseda.jp
  ieyuharu@ruri.waseda.jp).
  } 
\thanks{Yixuan Li is with the School of Computer Science and Engineering, Southeast University, China (e-mail: yixuanli@seu.edu.cn).}

\thanks{
Osamu Yoshie is the corresponding author. 
} 
}

\markboth{IEEE Trans. on MULTIMEDIA,~Vol.~XX, No.~XX, March~2024}%
{Shell \MakeLowercase{\textit{et al.}}: A Sample Article Using IEEEtran.cls for IEEE Journals}


\maketitle

\begin{abstract}

Large-scale text-to-image diffusion models have achieved unprecedented success in image generation and editing. However, extending this success to video editing remains challenging. Recent video editing efforts have adapted pretrained text-to-image models by adding temporal attention mechanisms to handle video tasks. Unfortunately, these methods continue to suffer from temporal inconsistency issues and high computational overheads. In this study, we propose FluencyVE, which is a simple yet effective one-shot video editing approach. FluencyVE integrates the linear time-series module, Mamba, into a video editing model based on pretrained Stable Diffusion models, replacing the temporal attention layer. This enables global frame-level attention while reducing the computational costs. In addition, we employ low-rank approximation matrices to replace the query and key weight matrices in the causal attention, and use a weighted averaging technique during training to update the attention scores. This approach significantly preserves the generative power of the text-to-image model while effectively reducing the computational burden. Experiments and analyses demonstrate promising results in editing various attributes, subjects, and locations in real-world videos. Our code is available at \url{https://github.com/CIMASA/FluencyVE}.

\end{abstract}

\begin{IEEEkeywords}
Diffusion models, video editing, one-shot, Mamba and fine-tuning.
\end{IEEEkeywords}


\section{Introduction}
\label{sec:intro}

\IEEEPARstart{D}{iffusion} -based models (DMs) have revolutionized text-to-image (T2I) generation, showcasing unprecedented capabilities in creating high-quality and diverse images \cite{ho2020denoising}. These models, such as Stable Diffusion (SD) \cite{rombach2022high} and DALL·E\cite{ramesh2021zero}, leverage the power of large-scale pretrained language models to generate content that is consistent with textual cues, vastly outperforming previous generative adversarial network (GAN) based models in terms of fidelity and diversity. Recent advancements, such as ControlNet \cite{zhang2023controlnet}, are compatible with pretrained SD models and enhance them through fine-tuning of the attention layers. These methods enable users to make precise modifications to image objects, background styles, and more, using textual descriptions.

\IEEEpubidadjcol

Diffusion-based T2I models excel in image generation and editing but face challenges in video editing. Text-driven video editing must address three key concerns: (1) semantic alignment, ensuring that the edited video accurately reflects the intended textual prompts; (2) spatial coherence, maintaining consistency between each modified frame and its corresponding frame in the original video; and (3) temporal continuity, ensuring smooth and seamless motion across frames for a cohesive viewing experience. Currently, video generation methods can be broadly categorized into two approaches: the first involves training text-to-video (T2V) diffusion models on large-scale text-video datasets, such as Imagen Video\cite{ho2022imagen} and MagicVideo\cite{zhou2022magicvideo} ; the second focuses on adapting existing T2I diffusion models for video generation. Owing to the difficulty in acquiring large-scale text-video datasets and the high computational cost of training T2V models, modifying and extending T2I models has become a more feasible option. While T2I models excel at capturing spatial features, they lack the ability to model temporal dimensions, which poses challenges in maintaining motion and temporal consistency in video generation. To address the challenge of temporal modeling in T2I models, Tune-A-Video\cite{wu2023tune} introduced a temporal attention layer, enabling efficient time-series modeling. By fine-tuning pretrained T2I models, it achieves one-shot video tuning for T2V generation, thereby eliminating the need for large-scale video datasets. Building upon this foundation, subsequent works such as CAMEL\cite{zhang2024camel} and SAVE\cite{karim2023save} have further refined the attention mechanism through various approaches, leading to improved video editing quality. Despite the significant success of these methods, they reduce the computational overhead by employing sparse attention, which compromises the global frame attention and can negatively impact the temporal consistency. In addition, excessive parameter fine-tuning of the pretrained T2I model can degrade its original generative performance while simultaneously increasing the computational costs.

In our study, we introduced the linear time-series model Mamba\cite{gu2023mamba} to enhance the global frame attention and temporal consistency. With a computational complexity of \(O(N)\), Mamba is more efficient than self-attention, allowing deeper model architectures with efficient memory utilization. In addition, we proposed a novel fine-tuning method known as Bypass Attention. Unlike traditional efficient fine-tuning methods and the LoRA\cite{hu2021lora} approach, our method takes advantage of low-rank approximation matrices of sizes \(k\times d\) and \(d\times k\) to replace the original \(W_{Q}\) and \(W_{K}\) matrices with the same size of \(d \times d\). This adjustment significantly reduces the tunable parameters while also lowering the overall computational overhead. The contributions of this study are summarized as follows:
\begin{itemize}
    \vspace{-3mm}
    \item We propose an efficient linear temporal-aware Mamba module for video tasks, enhancing the global frame attention with denser attention while increasing the network depth through stacking at a minimal computational cost.
    \item We introduce a novel fine-tuning method for casual attention, using low-rank approximation matrices to reduce the computational overhead and minimize the impact on the T2I model parameters.
    \item Experiments show that our model significantly improves the one-shot video editing training speed, delivering better editing results than other models.
\end{itemize}


\section{Related Work}
\subsection{Text-to-Image Model}

Advancements in GANs and diffusion models have significantly propelled text-to-image (T2I) generation \cite{8451971}. GAN-based T2I models \cite{reed2016generative} outperformed early methods like VQ-VAE \cite{van2017neural} through adversarial generator-discriminator structures but faced challenges such as unstable training and mode collapse. To address these issues, DALL·E \cite{ramesh2021zero} utilized Transformers \cite{vaswani2017attention}, showcasing the potential of attention mechanisms for complex text-image alignment. Diffusion models (DMs), such as DDPM \cite{ho2020denoising}, further improved stability and image quality through noise-learning processes. Latent Diffusion Models (LDM) \cite{rombach2022high} optimized diffusion in latent space, reducing computational costs while enabling high-resolution tasks like LinFusion’s 16K zero-shot image generation. Stable Diffusion (SD) \cite{rombach2022stable} enhanced text-image generation by integrating pretrained language models like CLIP \cite{Radford2021LearningTV}, enabling accurate semantics and compatibility with extensions like ControlNet \cite{zhang2023controlnet} and LoRA \cite{hu2021lora}. Recent innovations, such as IP-Adapter \cite{ye2023ip} and InstantID \cite{wang2024instantid}, demonstrate efficient and controllable generation by fine-tuning attention layers in SD, solidifying roles in SD in image and video editing.

\subsection{Video Editing Model}

Early video editing models relied on heuristic approaches and handcrafted algorithms, such as optical flow-based motion tracking and color-based background segmentation \cite{9257101,10420468,10345745}. Deep learning models, including CNNs and RNNs, introduced generative capabilities, with approaches like Vid2Vid \cite{wang2018vid2vid} and sequence models such as LSTMs and C3D \cite{tran2015learning}. Transformers, such as VideoBERT \cite{sun2019videobert} and TimeSformer \cite{bertasius2021timesformer}, further improved editing by leveraging attention mechanisms, while multimodal models like MMVID \cite{seo2022mmvid} extended capabilities by integrating text and audio inputs.

Diffusion models (DMs) transformed text-driven editing with fine-grained control. Techniques like Textual Inversion \cite{gal2022image}, Prompt-to-Prompt \cite{hertz2022prompt}, and Blended Diffusion \cite{avrahami2022blended} enabled localized edits, while ControlNet \cite{zhang2023adding} introduced external control signals. In video editing, methods like Tune-A-Video \cite{wu2023tune} and VideoP2P \cite{liu2024video} enhanced temporal consistency by employing cross-frame attention mechanisms and frame-guided denoising, mitigating issues such as flickering and drift. However, they still face challenges in long-range consistency and precise motion coherence. 

Training-free models, including Render-A-Video \cite{yang2023rerender}, Text2Video-Zero \cite{text2video-zero}, and TokenFlow \cite{tokenflow2023}, enable efficient editing by leveraging pretrained diffusion priors, though they encounter memory constraints and limitations in temporal perception. Recent advancements further address these limitations: FLATTEN \cite{cong2024flatten} employs optical flow-guided attention for improved frame alignment, RAVE \cite{kara2024rave} introduces randomized noise shuffling to enhance consistency and speed, Slicedit \cite{cohen2024slicedit} utilizes spatio-temporal slicing to maintain fine-grained edits across frames, and Factorized Diffusion Distillation \cite{singer2024video} improves video editing efficiency via factorized learning. These techniques enhance controllability but often struggle to balance semantic coherence, temporal consistency, and editing efficiency. As a result, improvements in one aspect frequently come at the cost of another. To address these challenges, we propose a linear-complexity sequence model to better capture temporal dynamics and achieve semantically consistent video generation over long sequences.

\subsection{State Space Models}

Early sequence models like RNNs effectively captured temporal dependencies but struggled with long sequences due to vanishing gradients and lack of parallelization \cite{sutskever2014sequence}. Transformers \cite{vaswani2017attention} resolved these issues with self-attention, enabling parallel processing and superior long-range modeling, though their quadratic complexity posed challenges for longer sequences. This limitation spurred the development of state space models (SSMs) \cite{gu2021combining}, which offer linear complexity and efficient handling of long sequences. The introduction of S4 \cite{gu2021efficiently} advanced SSMs by leveraging HiPPO \cite{gu2020hippo} projections to model long-range dependencies while maintaining linear scaling. Despite its strengths, S4 lacked selective attention for specific inputs. Mamba \cite{gu2023mamba} addressed this with selective scan algorithms and hardware-aware optimizations like parallel scanning and kernel fusion, enhancing computational efficiency and input focus. Mamba further improved flexibility for tasks requiring precise input control while preserving linear complexity. VisionMamba \cite{liu2024vmambavisualstatespace}, building on Mamba, applied bidirectional SSMs to dense visual prediction tasks, achieving superior memory efficiency and faster inference without relying on 2D priors. This allowed VisionMamba to outperform attention-based models like DeiT \cite{touvron2021training} in object detection and segmentation \cite{li2022exploring}. Extending VisionMamba, VideoMamba \cite{li2024videomamba} introduced dynamic spatiotemporal modeling for video understanding, excelling in short-term action recognition and long-term video comprehension, as demonstrated on datasets like Kinetics-400 \cite{kay2017kineticshumanactionvideo} and COIN \cite{tang2019coin}. VideoMamba’s efficiency, combined with its capability to handle multimodal tasks like video-text retrieval, positions it as a leading solution for comprehensive video understanding.

\subsection{Efficient Fine-Tuning}

Pre-trained models trained on large datasets extract both shallow and deep features, enabling transfer learning via fine-tuning. While early approaches adjusted all parameters, the growing size of models like GPT-4 \cite{achiam2023gpt} has made full fine-tuning impractical due to high costs and catastrophic forgetting. Selective fine-tuning methods, such as BitFit \cite{zaken2021bitfit} and LT-SFT \cite{ansell2021composable}, update only specific parameters but struggle with complex tasks. Additive fine-tuning, including Prefix Tuning \cite{li2021prefix} and adapter-based methods like IP-Adapter \cite{ye2023ip}, enhances task adaptation but increases computational costs. Parameter-efficient methods like LoRA \cite{hu2021lora} use low-rank decomposition to minimize updates and allow modular task switching. However, LoRA may degrade performance in video editing tasks with large parameter requirements. Optimizing these methods is essential to balance efficiency and complexity in such scenarios.


\section{Methodology}
Starting with a given video template and its associated text prompt, the task is to generate a new video based on the modified text prompt $P^*$. The key challenge lies in ensuring that the generated video not only maintains semantic consistency with $P^*$ but also preserves the motion continuity of the original video template. Notably, our approach leverages pre-trained text-to-image (T2I) models. 

\figmethod

\subsection{Preliminaries}

\label{subsec:Preliminaries}

\paragraph{Stable Diffusion Model.}
Our model is developed based on Stable Diffusion (SD) Model \cite{rombach2022high}. The SD Model is a variant of DDPM\cite{ho2020denoising}, and unlike DDPM, which restores the latent representation of the data from random noise through a process of gradually adding and removing noise, SD performs denoising in the latent space. Specifically, the forward process of SD maps the input image to the latent space by compressing it through an autoencoder\cite{van2017neural}, representing it as a latent variable $z_0$. In this latent space, the model gradually adds noise to $z_0$ through a predefined Markov chain, generating a sequence of increasingly noisy latent variables $z_t$. Each step of the transition process follows the Gaussian distribution:
\vspace{-1mm}
\begin{equation}
q(z_t | z_{t-1}) = \mathcal{N}(z_t; \sqrt{1 - \beta_t} z_{t-1}, \beta_t \mathbf{I}), \quad t = 1, \dots, T
\end{equation}
where $T$ is the total number of steps in the forward diffusion process. The parameter $\beta_t$ controls the noise strength at timestep $t$. In the generation phase, the model learns a reverse diffusion process, starting from random noise and gradually removing the noise to recover the latent representation, which is finally decoded into the target image. The reverse process is parameterized by a neural network, with the goal of predicting the denoising path given the noisy state:
\vspace{-1mm}
\begin{equation}
p_\theta(z_{t-1} | z_t) = \mathcal{N}(z_{t-1}; \mu_\theta(z_t, t), \Sigma_\theta(z_t, t))
\end{equation}

where $\mu_\theta$ and $\Sigma_\theta$ are the mean and covariance predicted by the model, guiding the denoising process at each step. For text-guided SD, the objective is expressed as:
\vspace{-1mm}
\begin{equation}
L = \mathbb{E}_{z, \epsilon \sim \mathcal{N}(0,1), t, c} \left[ \| \epsilon - \epsilon_\theta(z_t, t, c) \|_2^2 \right]
\end{equation}

where $c = \psi(\mathcal{P^*})$ is the embedding of the text condition $\mathcal{P^*}$, processed by the text-encoder CLIP ViT-H/14\cite{Radford2021LearningTV}.

\paragraph{Network Inflation.}
To transform a text-to-image (T2I) model into a text-to-video (T2V) model, the 2D U-Net architecture is expanded to capture both spatial and temporal information. First, the 2D convolutional layers are inflated into pseudo 3D convolutional layers by replacing the   $3 \times 3$ kernels with $1 \times 3 \times 3$ kernels, allowing the model to process video sequences. In addition, temporal self-attention layers are introduced to capture the relationships between frames using the formula:
\vspace{-1mm}
\begin{equation}
\mathrm{Attention}(Q, K, V) = \mathrm{Softmax}\left( \frac{Q K^T}{\sqrt{d}} \right) \cdot V
\end{equation}

where $Q$, $K$, and $V$ are derived from the video frame features. To improve the computational efficiency, a sparse causal attention mechanism is employed, computing attention only between the current frame $z_{v_i}$ and two previous frames $z_{v_1}$ and $z_{v_{i-1}}$:

\begin{equation}
\begin{aligned}
    Q &= W_Q z_{v_i}, \quad K = W_K [z_{v_1}, z_{v_{i-1}}], \\
    V &= W_V [z_{v_1}, z_{v_{i-1}}]
\end{aligned}
\end{equation}

This reduces the complexity to $\mathcal{O}(2mN^2)$, where \(m\) denotes the number of frames considered in the attention mechanism, ensuring efficient video generation with temporal consistency.

\paragraph{Mamba.}
Our model uses Mamba as the underlying framework, which is based on State Space Models (SSMs) and discretizes continuous systems to map a 1D sequence $x(t) \in \mathbb{R}$ to an output $y(t) \in \mathbb{R}$ via a hidden state $h(t) \in \mathbb{R}^N$. This is governed by the differential equations:
\vspace{-1mm}
\begin{equation}
    h'(t) = A h(t) + B x(t), \quad y(t) = C h(t)
\end{equation}
where $A \in \mathbb{R}^{N \times N}$ is the evolution matrix, and $B \in \mathbb{R}^{N \times 1}$, $C \in \mathbb{R}^{1 \times N}$ are the projection matrices. The zero-order hold method discretizes $A$ and $B$ as follows:
\vspace{-1mm}
\begin{equation}
    A_d = \exp(\Delta A), \quad B_d = (\Delta A)^{-1} (\exp(\Delta A) - I) \Delta B
\end{equation}

Following discretization, the system can be described as:
\vspace{-1mm}
\begin{equation}
    h_t = A_d h_{t-1} + B_d x_t, \quad y_t = C h_t
\end{equation}

Mamba introduces a selective scan mechanism (S6) to adaptively adjust $B$, $C$, and $\Delta$ adaptively based on the input. Finally, the output is computed via the convolution kernel $K$:
\vspace{-1mm}
\begin{equation}
    K = \left(CB, CAB, \dots, CA^{M-1}B\right)
\end{equation}

This approach combines discretized SSMs with global convolution, making it suitable for long-sequence tasks.

\figscan
\subsection{Temporal-Aware Mamba}
\label{subsec:Temporal-Aware Mamba}
Previous video editing methods often employed causal and temporal attention to process sequences. However, temporal attention, which focuses on the same spatial positions across frames, struggles to capture the global frame correlations, resulting in poor temporal continuity. Extending attention from keyframes to entire sequences could improve the performance, but the \(O(T^2)\) complexity of the self-attention makes the global attention impractical for long videos.

To address this challenge, we integrated the linear sequence processing model, Mamba \cite{gu2023mamba}, into our task. Drawing inspiration from VideoMamba \cite{li2024videomamba}, we followed the Spatial-First rule and developed four distinct scanning methods, as shown in Figure \ref{fig:scan}. If frames containing temporal and spatial information are treated as a sequence of tokens, the four scanning methods can be categorized as: (a) Spatial forward, Temporal forward; (b) Spatial forward, Temporal reverse; (c) Spatial reverse, Temporal forward; and (d) Spatial reverse, Temporal reverse. 

In contrast to the self-attention mechanism employed in Transformers, Mamba, as a linear sequence model that focuses on local dependencies, is more sensitive to neighboring tokens than to distant ones. This limitation reduces its effectiveness in distinguishing between frames in video sequences. To address this issue, we introduce a padding method with a trainable frame embedding. As shown in Fig. \ref{fig:padding}(a), for each video frame at time \( t \), represented as \( x_{t} \in \mathbb{R}^{H \times W} \), we pad the frame with a learnable embedding \( \theta_{\text{frame}} \), resulting in a padded frame \( x_{t}' \in \mathbb{R}^{(H+2) \times (W+2)} \). This padding enables Mamba to better differentiate between frames and improves its ability to learn the intra-frame feature distributions.
Next, as illustrated in Fig. \ref{fig:padding}(b), three different flip operations are applied to \( x_{t}' \), producing four variations (including the original). These inputs are processed through the same convolution, activation, and state-space model (SSM) pipeline:
\vspace{-1mm}
\begin{equation}
    z_{t}^{(i)} = \text{SSM}(\sigma(\text{Conv}(\text{flip}_i(x_{t}')))), \quad i = 0, 1, 2, 3,
\end{equation}

where \( \text{flip}_0 \) represents no flipping (the original input). After processing, we obtain the fusion feature \(z_{t}\) as shown below: 
\vspace{-1mm}
\begin{equation}
    z_{t,\text{final}} = z_{t,\text{flip}}^{(0)} + \sum_{i=1}^{3} \text{flip}_i^{-1}(z_{t}^{(i)}).
\end{equation}

where \( z_{t,\text{flip}}^{(0)} \) denotes the restoration of all outputs \(z_{t}^{(i)}\) to a form with the same temporal ordering as the original inputs.

\figpadding

\subsection{Bypass Attention}
\label{subsec:Bypass Attention}

The bypass attention mechanism is developed to provide a parameter-efficient alternative for the denoising network within the pretrained diffusion model. Specifically, it introduces a bypass network designed to compute attention maps with improved parameter efficiency. This bypass network generates a new attention map, denoted as \(A_\phi'\), which is subsequently integrated with the original attention map from the pretrained diffusion model, defined as \(A_\phi= Q W_QW_K^T K^T\). To ensure both efficiency and effectiveness, the bypass attention mechanism incorporates two key features: maintaining dimensional consistency between \(A_\phi'\) and \(A_\phi\), and minimizing the discrepancy between them prior to fine-tuning to facilitate optimization. To achieve these objectives, two low-rank approximations, \(W_Q' \in \mathbb{R}^{d \times k}\) and \(W_K' \in \mathbb{R}^{d \times k}\), are introduced as substitutes for the original projection matrices, \(W_Q \in \mathbb{R}^{d \times d}\) and \(W_K \in \mathbb{R}^{d \times d}\), respectively, where \(k < d\). By replacing \(W_Q'W_K'^T\) with \(W_QW_K^T\), the attention maps can be computed while preserving the same dimensionality:
\vspace{-1mm}
\begin{equation}
A_\phi'(K, Q) = Q W_Q'W_K'^T K^T
\end{equation}

During the fine-tuning process, only \(W_Q'\) and \(W_K'\) are updated, while all other parameters are inherited from the pretrained T2I diffusion model and remain fixed. The final attention map in the bypass attention module is computed as a weighted combination of the new and original attention maps:
\vspace{-1mm}
\begin{equation}
    A_{\phi_{\text{full}}} (K, Q) = (1 - \varphi) \times A_\phi' (K, Q) + \varphi \times A_\phi (K, Q)
\end{equation}

\figfinetune

In addition, we focus on the initialization of the bypass attention, Our main objective is to maintain the calculation pattern and results between \( A_\phi' \) and \( A_\phi \) before fine-tuning, ensuring that we optimize a similar function  with significantly higher parameter and memory efficiency. By mimicking the calculation pattern of  \( A_\phi \), the bypass attention can achieve higher video editing performance . We demonstrate that by properly initializing the parameters and dimensions for \( W_{Q}' \) and \( W_{K}' \), we can ensure that the Frobenius norm distance between \( W_{Q}' W_{K}'^T \) and \( W_{Q} W_{K}^T \) is a small error of high order. This indicates that we can utilize the \( A_\phi' \) to inherit a similar calculation pattern from \( A_\phi \). To support our analysis, we employ the Johnson-Lindenstrauss lemma and Eckart-Young-Mirsky’s theorem.

\textbf{\textit{Johnson-Lindenstrauss lemma}}: Let $R\in\mathbb{R}^{d\times k}$ be a matrix with $i.i.d$. entries from $\mathcal{N}(0,1/k)$, where $1\leq k\leq d$. For any $y$, $z \in \mathbb{R}^{d}$, we have:
\vspace{-1mm}
\begin{equation}
\Pr\left(\left\|zRR^Ty^T-zy^T\right\|\leq\epsilon\left\|zy^T\right\|\right)>1-2e^{-\left(\epsilon^2-\epsilon^3\right)k/4}
\vspace{-1mm}
\end{equation}

\textbf{\textit{Eckart-Young-Mirsky’s theorem}}: Let $A$ be a matrix of rank $r$ and $A_{k}$ be a matrix of rank $k$, where $k<r$. The best choice for $A_{k}$, which minimizes the distance to $A$ measured by the operator norm or Frobenius norm, is obtained by truncating the singular value decomposition of $A$ at the $k^{th}$ term:
\vspace{-1mm}
\begin{equation}
\begin{aligned}
||A-A_k||=\left\|A-\sum_{i=1}^ks_iu_iv_i^T\right\|=\min_{rank(A^{\prime})\leq k}\|A-A^{\prime}\|
\end{aligned}
\vspace{-1mm}
\end{equation}
By combining Johnson-Lindenstrauss lemma and Eckart-Young-Mirsky’s theorem, when we define $W_{Q}^{'}=\sum^{k}_{i=1}s_{i}u_{i}$ and $W_{K}^{'}=\sum^{k}_{i=1}v_{i}$, we can infer the following inequality:
\vspace{-1mm}
\begin{equation}
\begin{aligned}
\begin{aligned}&\Pr\left(\left\|W_Q^{\prime}W_K^{\prime T}-W_QW_K^T\right\|\leq\epsilon\left\|W_QW_K^T\right\|\right)\\&\geq\Pr\left(\left\|W_QRR^TW_K^T-W_QW_K^T\right\|\leq\epsilon\left\|W_QW_K^T\right\|\right)\\&\geq1-o(1)\end{aligned}
\end{aligned}
\vspace{-1mm}
\end{equation}

Here, the Frobenius norm distance between \( W_{Q}' W_{K}'^T \) and \( W_{Q} W_{K}^T \) is a small error of high order which suggests that the calculation result for \( A_\phi' \) is able to mimic \( A_\phi \) before fine-tuning.


\figgoodcase
\vspace{-2mm}
\section{Experiments}

\subsection{Implementation Details}

In our experiments, we base our implementation on latent DMs \cite{rombach2022high} and primarily utilize the publicly released weights from the SD model\footnote{\url{https://huggingface.co/CompVis/stable-diffusion-v1-4}}. From the input video, we extracted 32 uniformly distributed frames at a resolution of 512 × 512, and applied our method to fine-tune the models over 500 steps with a learning rate of $3 \times 10^{-5}$ and a batch size of 1. During inference, we employed the DDIM sampler \cite{song2020denoising} with 50 steps, alongside classifier-free guidance \cite{ho2022classifier} with a guidance scale of 12.5. All experiments were conducted on a single NVIDIA A800 GPU using PyTorch 2.0.0.

\textbf{Dataset.} We conducted comparative experiments on 53 videos from the LOVEU-TGVE competition\cite{wu2023cvpr2023textguided}, comprising 16 videos from the DAVIS dataset\cite{perazzi2016benchmark} and 37 from Videvo. Each video was uniformly sampled into 32 frames at a resolution of 480 × 480. In addition, each video was paired with a ground-truth caption and four creative text prompts designed for tasks such as object manipulation, background modification, style transformation, and multiple changes.

\begin{table*}[ht]
    \caption{\textbf{Quantitative comparison} with other video editing methods. $\uparrow$ indicates that a higher value is advantageous.}
    \centering
    \resizebox{\textwidth}{!}{%
    \begin{tabular}{lcccccc}
        \toprule
        \multirow{2}{*}{\textbf{Method}} & \multicolumn{2}{c}{\textbf{Frame Consistency}} & \multicolumn{2}{c}{\textbf{Textual Alignment}} & \multirow{2}{*}{\textbf{Pick Score}} \\
        & \textbf{CLIP Score} $\uparrow$ & \textbf{User Preference} $\uparrow$ & \textbf{CLIP Score} $\uparrow$ & \textbf{User Preference} $\uparrow$ & \\
        \midrule
        Tune-A-Video & 92.923 & 19.9 & 27.675 & 14.9 & 20.658 \\
        CAMEL & 93.332 & 20.8 & 26.645 & 18.6 & 20.136 \\
        SAVE & 94.846 & 18.2 & 28.299 & 17.6 & 20.695 \\
        VidToMe & 95.516 & 18.4 & 28.844 & 22.3 & 20.665 \\
        Slicedit & 95.602 & 20.3 & 24.183 & 10.2 & 20.342 \\
        \midrule
        \textbf{Ours} & \textbf{96.465} & \textbf{22.7} & \textbf{29.419} & \textbf{26.6} & \textbf{20.744} \\
        \bottomrule
    \end{tabular}%
    }
    \label{tab:metric}
    \vspace{-2mm}
\end{table*}

\subsection{Main Results}

We demonstrate the outstanding editing performance of our method in Fig. \ref{fig:goodcase}. Specifically, it includes two video cases, with the Source Prompts being ``A cat in the grass in the sun.'' and ``A jeep car is moving on the road.''

\textbf{Editing Background.}: As shown in Figure \ref{fig:goodcase}, rows 2 and 6, FluencyVE demonstrated strong performance in background editing. It is evident that ``grass" and ``road" were accurately transformed into ``beach" and ``snow", respectively. This indicates that FluencyVE not only retains the generative capabilities of the SD model but also enhances attention to global frame details. As shown in the second row of Figure 6, when converting ``grass" into ``beach," the model also adjusted the video to a more suitable color tone.

\textbf{Editing Style.}: Attention to global spatiotemporal features is especially critical for video style editing. In the 8th row of Figure \ref{fig:goodcase}, we added a ``cartoon style" to the input video. It can be observed that our approach successfully transforms all frames into the target style without altering the semantics of the original video.

\textbf{Replacing Subjects.}: Rows 3, 4, 7, and 8 in Fig. \ref{fig:goodcase} illustrate the effectiveness of FluencyVE in video object replacement tasks. In rows 3 and 4, FluencyVE successfully replaced a cat with a dog and a lion, while in Rows 7 and 8, a jeep was convincingly replaced with a sports car and an AE86. These results demonstrate that our edits maintain semantic consistency with the prompts while remaining faithful to the original video. Furthermore, FluencyVE supports the ability to edit multiple attributes within a video. For example, in Row 4, FluencyVE not only replaced the cat with a lion but also added butterflies. Similarly, in row 8, the jeep was replaced with an AE86, and the overall video style was transformed into a cartoon aesthetic.

\figcomparison

\subsection{Comparisons}

\textbf{Baselines.} We evaluated our method against four state-of-the-art(SOTA) video editing approaches: \textit{(1)Tune-A-Video (TAV)} \cite{wu2023tune}: A widely recognized SOTA method in video editing, serving as a conventional baseline for related works.  \textit{(2)CAMEL} \cite{zhang2024camel}: An extension of TAV that introduces causal motion enhancement. \textit{(3)SAVE} \cite{karim2023save}: Builds upon TAV by incorporating spectral-shift-aware adaptation. \textit{(4)VidToMe} \cite{li2024vidtome}: A zero-shot video editing method based on self-attention tokens merging. \textit{(5)Slicedit} \cite{cohen2024slicedit}: A zero-shot video editing method using Spatio-Temporal Slices.

\begin{table}[htbp]
\caption{Comparison of parameter size, memory usage and Inference time across different methods.}
\begin{tabular}{lccc}
\hline
\textbf{Method} & \textbf{Para./M} $\downarrow$ & \textbf{Memory} $\downarrow$ & \textbf{Inf./S} $\downarrow$ \\ \hline
Tune-A-Video & 73.68 & 100\% & 50\\
Tune-A-Video+LoRA & 61.73 & 92\% & 42\\
CAMEL & 79.35 & 61\% & 56\\
SAVE  & 79.77 & 88\% & 75\\
VidToMe  & 87.51 & 91\% & 66\\
Slicedit  & 81.42 & 97\% & 149\\\hline
\textbf{FluencyVE w/o bypass attention} & \textbf{75.96} & \textbf{100\%} & \textbf{64} \\
\textbf{FluencyVE} & \textbf{4.80} & \textbf{59\%} & \textbf{29} \\ \hline
\end{tabular}
\label{tab:FluencyVE}
\end{table}

\textbf{Qualitative Results.}
The results of different methods on the editing subjects are shown in Fig. \ref{fig:comparison}. It can be observed that TAV \cite{wu2023tune} achieved relatively better temporal consistency through its temporal modeling and one-shot tuning. However, the poses and motions in the edited video were not faithful to the original video. In addition, owing to excessive parameter adjustments, some frames became blurred, which compromised part of the performance of the SD model. While VidToMe\cite{li2024vidtome} and Slicedit\cite{cohen2024slicedit}, as zero-shot methods, remain faithful to the original video and ensure temporal consistency, their editing results fail to align with the intended semantics. For instance, VidToMe replaces the trackside marker with old tires suitable for an off-road setting, yet retains the number three, which breaks semantic coherence. Slicedit merely adjusts the video tone without altering key scene elements, resulting in semantically irrelevant edits. SAVE \cite{karim2023save} improved the fine-tuning of the SD model to ensure frame quality, but it did not enhance the temporal consistency, leading to flickering and disappearing frames between shots. CAMEL \cite{zhang2024camel} strengthened attention to motion, but some frames still failed to maintain temporal consistency. For example, certain frames in the muddy off-road track exhibit unnaturally bright regions lacking road textures. By contrast, our method enhances global frame attention, ensuring that the generated video is not only faithful to the original but also exhibits a high degree of temporal consistency and motion continuity across frames. As shown in Fig. \ref{fig:comparison}, our model successfully completes the transformation from car to truck, while accurately preserving the vehicle's original orientation and generating a background that best matches the intended semantics.

\figabaa

\textbf{Quantitative Results.}
We evaluated our method in comparison to baseline models using both automatic metrics and a user study, with the results for the frame consistency and textual faithfulness presented in Tab. \ref{tab:metric}.
\textbf{Automatic Metrics.} We note that there is no universally accepted evaluation standard for video editing. Therefore, we opted to use the three metrics provided by the LOVEU-TGVE competition\cite{wu2023cvpr2023textguided} as our evaluation criteria: (i) \textit{CLIP Score text}: This metric measures the alignment between video frames and a text prompt using the average cosine similarity of their embeddings obtained from a pretrained CLIP ViT-L/14 model\cite{Radford2021LearningTV}. (ii) \textit{CLIP Score frame}: Evaluates the frame consistency among video frames by computing the average cosine similarity between frame embeddings, ignoring self-similarity, to gauge the internal coherence of the video\cite{hessel2022clipscorereferencefreeevaluationmetric}. (iii) \textit{Pick Score}: This metric leverages a CLIP model trained on human preferences to evaluate the alignment between video frames and a given prompt from a human perspective\cite{kirstain2023pickapicopendatasetuser}. 
\textbf{User Study} Aligned with the two key objectives of video editing, we asked 150 users to select: a) the video with higher editing fidelity, and b) the one with better temporal consistency. Each user evaluated 30 randomly selected videos, and the final score was based on the percentage of users preferring each method.

\begin{table*}[ht]
    \caption{Quantitative ablation study for Mamba modules. $\uparrow$ indicates that a higher value is advantageous.}
    \label{tab:aba_mamba}
    \centering
        \begin{tabular}{c|cc|cc|ccc}
            \toprule
            Model & w/o Mamba & w/ Mamba& w/o Padding & w/ Padding & Depth=1 & Depth=2 & Depth=4 \\
            \midrule
            \textbf{Frame Consistency$\uparrow$} & 93.561 &\textbf{96.465} &94.109&\textbf{96.465}&95.277&\textbf{96.465}&94.969\\
            \textbf{Textual Alignment $\uparrow$} &28.415& \textbf{29.419}&28.762 &\textbf{29.419} &29.138&\textbf{29.419}&28.791\\
            \bottomrule
            
        \end{tabular}%
\end{table*}

\subsection{Ablation Study}
\label{subsec:Ablation Study}

\textbf{Padding Strategy} There are several possible means of differentiating between different time frames explicitly. One simple padding approach is to insert fixed tokens between the token sequences of each frame. Alternatively, as in our approach, padding can be applied by explicitly adding identical embeddings to each frame, where these embeddings have independent parameters. As shown in Fig. \ref{fig:aba_1}, in the case of replacing ``woman" with ``polar bear," omitting the padding method led to a noticeable disruption in the temporal consistency. As shown in the second row, the spatial position and movement of the polar bear shifted abruptly between frames. With simple padding, the temporal consistency improved; however, it lacked attention to the frame content, resulting in instances in which the polar bear’s face appeared in odd locations within some frames. Our padding strategy not only explicitly enhances the temporal awareness of the model but also strengthens the intra-frame attention, achieving optimal editing outcomes.

\textbf{Neural Network Depth} The introduction of time-aware Mamba modules significantly reduces the computational cost while allowing the network to increase the depth by stacking. Such stacking does not significantly affect the temporal consistency, but excessive stacking can affect the semantic consistency and increase the risk of overfitting. As shown in Fig. \ref{fig:aba_2}, Without stacking depth, while the Eiffel Tower was successfully replaced with the Canadian National Tower, the background was not effectively altered. With a stacking Mamba block depth of 2, both the Eiffel Tower and the background were successfully replaced, achieving a high level of semantic consistency. However, at a stacking depth of 4, both the Eiffel Tower and the background replacements failed, indicating significant model overfitting. Extensive experiments reveal that the best editing performance was achieved when the number of Mamba blocks was set to 2. When the number exceeded 4, the model's performance started to degrade significantly. 

\begin{table*}[t]
    \caption{Quantitative ablation study for Bypass Attention modules. $\uparrow$ indicates that a higher value is advantageous.}
    \centering
    \begin{tabular}{c|cc|cc|ccc}
        \toprule
        \textbf{Model} & \textbf{Frame Consistency$\uparrow$}& \textbf{Textual Alignment$\uparrow$}& \textbf{Para./M}$\downarrow$ &\textbf{Memory} $\downarrow$ & \textbf{Inf./S} $\downarrow$\\
        \midrule
        w/o Bypass attention & \textbf{96.525}& 29.172 & 75.96 & 100\% &64\\
        w/ Bypass attention & 96.465& \textbf{29.419} & \textbf{4.8} & \textbf{59}\% &\textbf{29}\\
        \bottomrule
        
    \end{tabular}%
    \label{tab:aba_by}
    
\end{table*}

For the Temporal-Aware Mamba module, Table \ref{tab:aba_mamba} provides strong empirical evidence of its positive impact on both frame consistency and textual alignment. Additionally, the results validate the effectiveness of the padding strategy and justify the choice of setting the Mamba network depth to 2.

\figabab

\textbf{Fine-Tuning Operation} Selecting the dimension  \(k\times d\) for the low-rank approximation matrix in our fine-tuning method is particularly crucial. A value of  \(k\) that is too small will lead to significant information loss and model degradation, while a value that is too large will substantially increase the training cost, defeating the purpose of fine-tuning. As shown in Fig. \ref{fig:aba_3}, in the style editing case, setting the approximate matrix dimension to 4 led to significant loss of prior information in our Bypass Attention, resulting in nearly no visible changes in the edited video. When the dimension was set to 8, the editing effect improved significantly; as shown in the fourth row, both the rabbit and the watermelon were converted into a cartoon style. At a value of 12, the best editing results were achieved—not only was the style correctly transformed, but the model also added more details to the rabbit and watermelon and refined the background to better align with semantic cues. After extensive testing, we found that when  \(k\) = 12, the model achieves satisfactory performance with a relatively fast convergence rate. On the other hand, we experimented with training from a randomly initialized low-rank approximation matrix, but as seen in the 5th row of Fig. \ref{fig:aba_3}, the results were suboptimal. This demonstrates that our initialization method is more robust.

According to Table \ref{tab:aba_by}, Bypass Attention fine-tuning method enables the pre-trained Stable Diffusion model to fully realize its potential in video editing tasks. It achieves a notable improvement of 0.241 in Textual Alignment while incurring only a minimal decline of 0.06 in Frame Consistency. Moreover, in terms of model efficiency, Bypass Attention significantly reduces the parameter count to just 6\% of the original, effectively doubles the inference speed, and substantially reduces memory consumption, demonstrating its remarkable effectiveness in optimizing computational efficiency without compromising performance.

\figabac

\section{Conclusion and Discussion}
We have proposed a novel and efficient text-driven video editing framework that integrates the linear time-series Mamba module and a refined scanning strategy to enhance global frame attention via denser attention mechanisms. This design substantially improves temporal consistency and motion continuity in video editing tasks. To mitigate performance degradation and computational overhead from extensive parameter tuning in T2I-based models, we introduce Bypass Attention, which replaces the Query and Key matrices with low-rank approximations, effectively reducing computational cost while preserving generative capacity. Extensive experiments validate the superior performance and training efficiency of our method. Although the approach is not training-free and still requires fine-tuning, future work will explore adapter-based designs to improve compatibility and further reduce training cost.



\bibliographystyle{IEEEtran}
\bibliography{ref}


\vfill
\end{document}